\documentclass[journal]{IEEEtran}
\usepackage{amsmath,amsfonts}
\usepackage{algorithm}
\usepackage{algpseudocodex}
\usepackage{array}
\usepackage[caption=false,font=normalsize,labelfont=sf,textfont=sf]{subfig}
\usepackage{textcomp}
\usepackage{stfloats}
\usepackage{url}
\usepackage{verbatim}
\usepackage{graphicx}
\usepackage{cite}

\usepackage{hyperref}
\usepackage{multirow}
\usepackage{amsthm}
\usepackage{amssymb}
\usepackage{xcolor}
\usepackage{overpic}
\usepackage{pifont}
\usepackage{booktabs}
\usepackage{diagbox}
\usepackage[switch]{lineno}

\allowdisplaybreaks

\newcommand{\cmark}{\ding{51}}
\newcommand{\xmark}{\ding{55}}

\newcommand{\eg}{\emph{e.g.}}
\newcommand{\etc}{\emph{etc}}
\newcommand{\etal}{\emph{et al.}}
\newcommand{\bc}{\mathbf{c}}
\newcommand{\bs}{\mathbf{s}}
\newcommand{\bh}{\mathbf{h}}
\newcommand{\bz}{\mathbf{z}}
\newcommand{\bI}{\mathbf{I}}

\newcommand{\br}{\mathbf{r}}
\newcommand{\be}{\mathbf{e}}
\newcommand{\bpi}{\boldsymbol{\pi}}
\newcommand{\bzero}{\mathbf{0}}
\newcommand{\bone}{\mathbf{1}}

\begin{document}

\title{Efficient Halftoning via Deep Reinforcement Learning}

\author{Haitian~Jiang,
  Dongliang~Xiong,
  Xiaowen~Jiang,
  Li~Ding,
  Liang~Chen,
  and~Kai~Huang
  \thanks{Manuscript received xx xx xxxx; revised xx xx xxxx. This work was supported by the National Key R\&D Program of China (2021YFB2206200). \emph{(Corresponding author: Kai Huang.)}}
  \thanks{Haitian Jiang, Dongliang Xiong, Xiaowen Jiang and Kai Huang are with the Institute of VLSI Design, Zhejiang University, Hangzhou, China (e-mail: jianghaitian@zju.edu.cn, xiongdl@zju.edu.cn, xiaowen\_jiang@zju.edu.cn, huangk@zju.edu.cn).}
  \thanks{Li Ding and Liang Chen are with Apex Microelectronics Co.,Ltd., Zhuhai, China (e-mail: ding@apexmic.com.cn, liang.chen@apexmic.com.cn).}
}

\markboth{IEEE Transactions on Image Processing,~Vol.~xx, No.~x, xxx~xxxx}%
{Jiang \MakeLowercase{\textit{et al.}}: Efficient Halftoning via Deep Reinforcement Learning}

\IEEEpubid{0000--0000/00\$00.00~\copyright~2023 IEEE}

\maketitle

\begin{abstract}
  Halftoning aims to reproduce a continuous-tone image with pixels whose intensities are constrained to two discrete levels. This technique has been deployed on every printer, and the majority of them adopt fast methods (e.g., ordered dithering, error diffusion) that fail to render structural details, which determine halftone's quality. Other prior methods of pursuing visual pleasure by searching for the optimal halftone solution, on the contrary, suffer from their high computational cost. In this paper, we propose a fast and structure-aware halftoning method via a data-driven approach. Specifically, we formulate halftoning as a reinforcement learning problem, in which each binary pixel's value is regarded as an action chosen by a virtual agent with a shared fully convolutional neural network (CNN) policy. In the offline phase, an effective gradient estimator is utilized to train the agents in producing high-quality halftones in one action step. Then, halftones can be generated online by one fast CNN inference. Besides, we propose a novel anisotropy suppressing loss function, which brings the desirable blue-noise property. Finally, we find that optimizing SSIM could result in holes in flat areas, which can be avoided by weighting the metric with the contone's contrast map. Experiments show that our framework can effectively train a light-weight CNN, which is 15x faster than previous structure-aware methods, to generate blue-noise halftones with satisfactory visual quality. We also present a prototype of deep multitoning to demonstrate the extensibility of our method.
\end{abstract}


\section{Introduction}
\label{sec:intro}
\IEEEPARstart{D}{igital} halftoning is the technique that converts continuous-tone images into images whose pixel values are limited to two discrete levels (black and white), which is the limitation of some rendering devices like printers.
Thanks to the low-pass filtering property of our human visual system (HVS), a halftone image can be perceived as its contone counterpart from a sufficient distance.
According to the dot clustering style, halftone images can be classified into clustered-dot and dispersed-dot.
Besides, there are two types of halftone textures: periodic and aperiodic \cite{lau2008}.
In this paper, we focus on halftones with \emph{dispersed-dot} and \emph{aperiodic} textures, which has been termed the blue-noise property \cite{ulichney1988}, since this kind of pattern gives the best visual pleasure.

In halftoning research, image quality and processing efficiency are of paramount concern.
With that in mind, we first briefly review three kinds of ``traditional'' halftoning techniques:
(1) Ordered dithering (screening) methods \cite{bayer1973,sullivan1991,ulichney1993,allebach1996} periodically split the contone image into tiles and compare them with a dither array, which was designed or generated in advance.
These methods are the fastest due to the high parallelism and less computation, but the quality of the resultant halftone is usually unsatisfactory.
(2) Error diffusion methods \cite{floyd1976,ulichney1988,ostromoukhov2001,li2004,chang2009,li2010,fung2016,hu2016,mao2018} quantize pixels in sequence.
At each step, the quantization error is diffused to adjacent unprocessed pixels with the predefined or input-dependent weights.
In general, error diffusion yields better halftones than ordered dithering, although the sequential processing style makes it difficult to compute in parallel and may introduce unpleasant artifacts.
(3) Search-based methods \cite{analoui1992,pang2008,arpitam2012} regard halftoning as an optimization problem.
They first design a halftone quality metric, which usually explicitly considers the HVS model \cite{nasanen1984}, then optimize it with heuristic algorithms like greedy search \cite{analoui1992} or simulated annealing \cite{pang2008}.
Search-based methods produce the best halftone quality out of the three types because they can directly optimize an elaborate halftone assessing metric (\eg, structural similarity \cite{wang2004}).
However, the computational cost of the searching is usually very expensive.

\IEEEpubidadjcol

In the wave of deep learning, inverse halftoning has greatly benefitted from the data-driven methodology \cite{kim2018,xia2019}.
However, there are much fewer discussions \cite{kim2018,guo2020,xia2021,choi2022} with respect to the deep ``forward halftoning''.
In this work, we propose a new data-driven halftoning method that can generate stochastic halftone images with structural details while remaining efficient.
Instead of searching just-in-time for an optimal solution of this high-dimensional combinatorial optimization problem, we train a light-weight fully convolutional neural network (CNN, with parameters $\theta$) in advance by optimizing the \emph{expectation} of the halftone metric.
At runtime, a halftone image can be generated quickly by one CNN inference.

\begin{table*}[t]
  \renewcommand{\arraystretch}{1.3}
  \caption{Comparison of halftoning formulations. $\bh$ and $\bc$ denote halftone and contone images, respectively.\label{tab:compare_formulation}}
  \centering
  \begin{tabular}{c|cc|ccc}
    \toprule
    Formulation & Offline Phase                                                                                                                                      & Online Phase & Parallelism & Metric Optimizing \\
    \midrule \midrule
    Ordered Dithering
                & Design/generate dither array
                & Pixel-wise thresholding with the array
                & full                                                                                                                                               & \xmark                                         \\
    Error Diffusion
                & Design/generate diffusion weights
                & Diffuse binarizing error pixel-by-pixel
                & limited                                                                                                                                            & \xmark                                         \\
    Search-based
                & -
                & $\bh = \bh^* = \mathop{\arg\min}\limits_{\bh} \text{Metric}(\bh, \bc)$
                & limited                                                                                                                                            & \cmark                                         \\
    DRL-based (ours)
                & $\theta^* = \mathop{\arg\min}\limits_{\theta} \mathbb{E}_{\bh \sim \text{Bernoulli}(\text{CNN}(\bc;\theta))} \left[\text{Metric}(\bh, \bc)\right]$
                & $\bh = \text{Thresholding}\left(\text{CNN}\left(\bc;\theta^*\right),0.5\right)$
                & full                                                                                                                                               & \cmark                                         \\
    \bottomrule
  \end{tabular}
\end{table*}

In training such a halftoning CNN, we find that the key difficulty of the learning lies in the final thresholding operation, whose derivative is zero almost everywhere that hinders the back-propagation of gradients.
In addition, there is no naturally existing ground-truth halftones available for learning.
Rather than learning from a prepared halftone dataset \cite{guo2020,choi2022} or using the empirical straight-through estimator (STE) \cite{bengio2013} and the binarization loss as in \cite{xia2021}, we formulate halftoning as a one-step multi-agent reinforcement learning (MARL) problem.
Specifically, each binary pixel in the output halftone is regarded as a stochastic action chosen by a virtual agent at the corresponding position with the shared CNN.
To make it possible to render aperiodic halftone patterns, we input the continuous-tone image with a Gaussian noise map \cite{xia2021} to the fully CNN followed by a pixel-wise sigmoid layer.
A tailored effective policy gradient estimator is proposed according to the computational characteristics of halftone metrics.
The differences between our deep reinforcement learning (DRL) based formulation and previous methods are summarized in Table~\ref{tab:compare_formulation}.
Moreover, in order to achieve the desirable blue-noise property, we design a novel loss function that suppresses the anisotropy of constant grayscale images' halftones in the frequency domain.
Last but not least, we find that the structural similarity index measure (SSIM) \cite{wang2004} used in \cite{pang2008} is actually deficient in optimizing halftone, resulting holes in flat areas with tones near 0/1.
To address this problem, we suggest optimizing the contrast-weighted SSIM (CSSIM), which weights the SSIM map with the contrast map of the continuous-tone image.

Extensive experimental results show the significance of our method.
We demonstrate that the proposed framework can enable a standard and light-weight CNN model, which is fully parallelizable and can be easily accelerated by modern deep learning HW/SW stacks (15x faster than prior structure-aware halftoning methods), to generate halftones with high quality.
The trained model achieves the best SSIM/CSSIM scores compared to existing methods, and the generated halftones can well preserve structural details and possess the blue-noise property.
Besides, the training converges quickly (about half a day on one consumer GPU).
The extensibility of our method is demonstrated by an example of extending to multitoning, in which the quantization level number is more than two.

Our contributions are summarized as follows:
\begin{itemize}
  \item We propose an efficient halftoning method via deep reinforcement learning. A fully convolutional neural network is trained to generate halftone images with structural details. The training converges fast thanks to the gradient estimator tailored according to the computational property of halftone assessing metrics.
  \item We propose the anisotropy suppressing loss function, which leads to the desirable blue-noise property.
  \item We point out that the SSIM metric has drawbacks in assessing halftones. We suggest optimizing CSSIM that weights the SSIM reward map with local contrast values of the continuous-tone image.
  \item Our method may facilitate more situations similar to the basic binary halftoning. A demonstration of extending to multitoning is shown in this paper.
\end{itemize}

The foundation of this work was reported in our preliminary study \cite{jiang2022}, named HALFTONERS.
On that basis, in the present paper, we provide three significant improvements:
First, we discuss the necessity of our DRL-based halftoning solution, establish connections to existing works \cite{furata2020,analoui1992,foerster2018,titsias2015}, and elaborate on how we design the framework in a step-by-step manner.
Second, we point out a fatal flaw in an existing halftone metric and solve it with a new weighting scheme.
Finally, we add more extensive experimental results, including component analyses, hyperparameter analyses, more comparisons \cite{chang2009,fung2016,hu2016,pang2008,choi2022}, and a multitoning prototype demonstrating our framework's extensibility.

\section{Preliminaries: DRL for Image Processing}
\label{sec:rl4ip}

In this section, we first discuss why we choose DRL for halftoning.
Then, a brief introduction of existing pixel-level DRL works for image processing is given.

\subsection{Why DRL?}
\begin{figure}[t]
  \centering
  \includegraphics[width=0.8\linewidth]{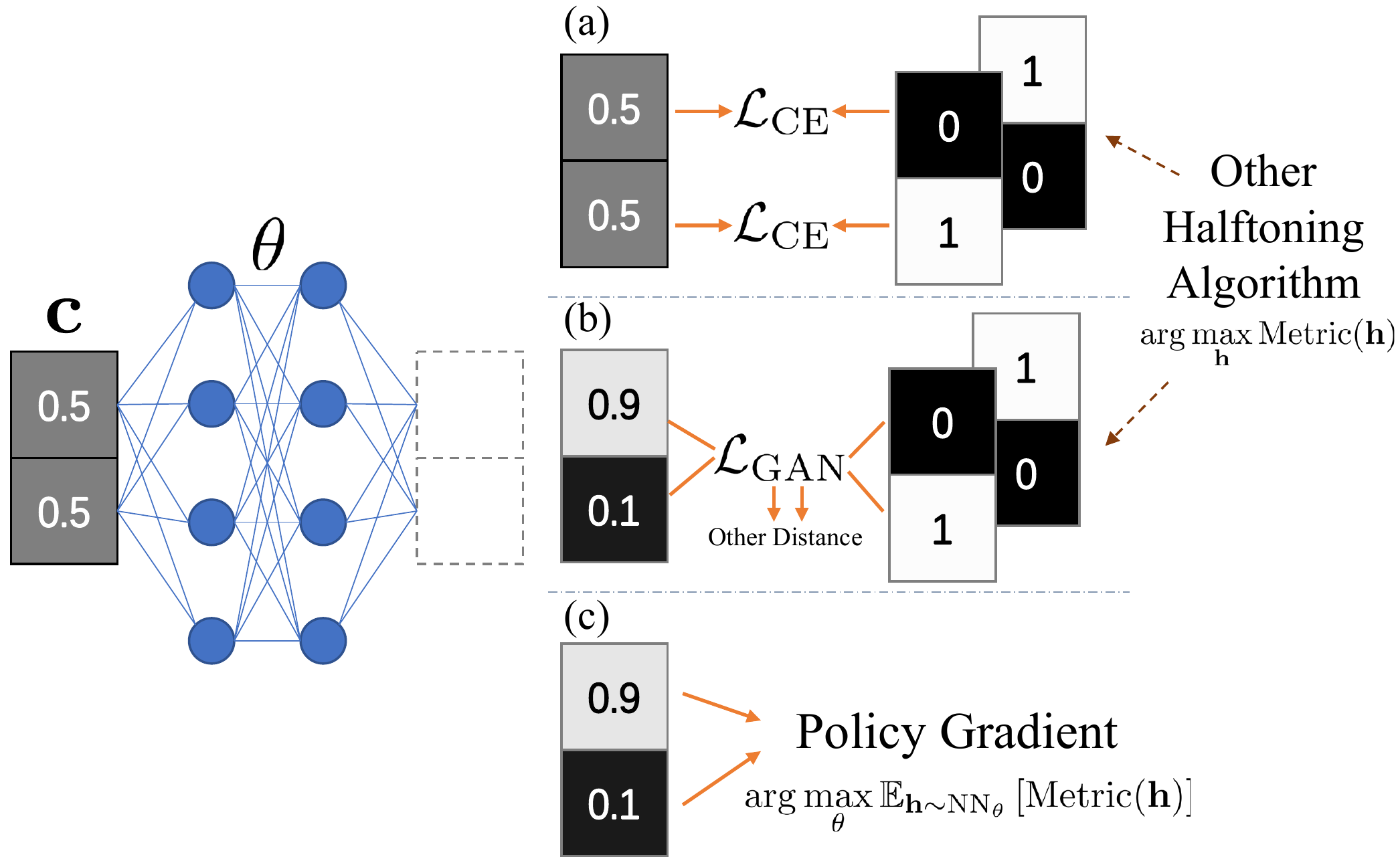}
  \caption{An illustration of adapting three common learning paradigms to halftoning. Here we dither a continuous-tone ``image'': [0.5;0.5], whose optimal halftones are [0;1] and [1;0]. (a) Supervised learning. Simply adopting pixel-wise distance loss like the cross-entropy $\mathcal{L}_\text{CE}$ fails to converge to the two discrete ends. (b) Unsupervised learning. GANs minimize another distance between the generated distribution and a pre-prepared target dataset. (c) Reinforcement learning. The policy network is directly trained to maximize the expectation of the target halftone metric.}
  \label{fig:SL_UL_RL}
\end{figure}

Supervised learning (SL), unsupervised learning (UL), and reinforcement learning (RL) are three common paradigms in deep learning.
Most image processing tasks that benefit from the data-driven methodology have adopted the former two, learning from a paired or unpaired image dataset.
In this work, we investigate how to train an efficient halftoning NN, which can dither an image by one forward computation.
However, we find that the common SL and UL are undesirable here:
(1) Preparing the ``ground truth'' halftones can be costly or even infeasible.
It's extremely expensive to obtain the optimal halftone image due to its NP-hard nature.
Existing search-based methods can only optimize certain metrics \cite{analoui1992} or utilize meta-heuristic search methods \cite{pang2008} that needs to be tuned per-instance \cite{xia2021}.
(2) It is non-trivial to transfer the (sub-)optimality from the halftone dataset to the neural network.
Halftoning is an one-to-many mapping problem in essence \cite{xia2021}.
Directly using a pixel-wise distance loss like the cross-entropy to optimize this ill-posed problem will actually learn an average image \cite{dahl2017,menon2020}, which violates halftone's discreteness constraint\footnote{Theoretically, autoregressive models \cite{dahl2017} can tackle this problem by modeling the dependency between discrete pixels. However, such models require a large amount of runtime \cite{choi2022} due to their iterative processing nature.}.
The generative adversarial network (GAN), which is a prevalent framework of UL, learns from a pre-prepared target halftone dataset \cite{kim2018,guo2020,choi2022}.
The discriminators in GANs are trained to measure another distance (\eg, the Pearson $\chi^2$ divergence used in LSGAN \cite{mao2017,choi2022}), which indirectly points to the ultimate halftone metric.
Elaborate designs are also needed to stabilize the training \cite{choi2022}.
Fig.~\ref{fig:SL_UL_RL}(a) and (b) illustrate these two situations.

Another approach to learn a dither network is to relax the discrete constraint \cite{yoo2020} and add a binarizing penalty loss which pushes the output values to their nearest discrete ends, as \cite{xia2021} did.
However, such a greedy binarizing rule could damage the optimization from a global perspective.

By contrast, we find that RL is a more appropriate paradigm for halftoning, especially the DRL combined with the powerful deep policy network.
In DRL, parameterized \emph{agent}(s) learns to maximize the accumulated \emph{reward} by interacting with the \emph{environment} \cite{sutton2018}.
The \emph{actions} that agents perform can be discrete, and the expected target metric (reward) can be directly and explicitly optimized by the policy gradient (see Fig.~\ref{fig:SL_UL_RL}(c)).
Besides, no additional ground-truth dataset is needed here.

\subsection{Pixel-level DRL}
There have recently been works that use DRL to solve image processing problems \cite{hu2018,furata2020,wang2022}.
While some methods learn to execute global actions on the entire image \cite{hu2018,li2018}, we are chiefly interested in the one whose actions are formulated at the pixel level: pixelRL, which was proposed by Furata \etal \cite{furata2020}.
Herein, we make a brief introduction.

In pixelRL, agents iteratively improve an image by performing actions.
Specifically, each pixel has a virtual agent, whose stochastic policy is denoted as $\text{Pr}(h_a^{(t)}|\bs_a^{(t)}) = \pi_a(h_a^{(t)}|\bs_a^{(t)})$, where $\bs_a^{(t)}$ and $h_a^{(t)}$ are the state and action of the $a$-th agent at the time step $t$, respectively.
There are $N$ agents ($\bpi=\{\pi_1,\ldots,\pi_N\}$) in total, which is also the number of pixels.
All agents share one fully convolutional network (FCN) instead of $N$ individual networks, so the learning can be computationally practical.
In the beginning $t=0$, the state $\bs^{(0)}$ is set to an initial solution with poor quality (\eg, the original noisy image in the image denoising task).
At each time step $t$, agents learn to take actions $\bh^{(t)} = \{h_1^{(t)},\ldots,h_N^{(t)}\}$ from a pre-defined toolbox $\mathcal{H}$ consisting of local operators, such as low-pass filtering and pixel value plus one, to improve the current $\bs^{(t)}$.
Then, the agents obtain the next state $\bs^{(t+1)}$ and rewards $\mathbf{r}^{(t)}=\{r_1^{(t)},\ldots,r_N^{(t)}\}$ from the environment.
The ultimate goal of the training is to maximize the expected cumulative rewards $\mathbf{R}^{(t)}=\sum_{t}^\infty \gamma^t \mathbf{r}^{(t)}$ that agents receive in the long run:
\begin{equation}
  J(\theta)
  =\mathbb{E}_{\bpi_\theta} \left[\frac{1}{N} \sum_{a=1}^N R_a^{(0)}\right]
  =\mathbb{E}_{\bpi_\theta} \left[\frac{1}{N} \sum_{a=1}^N \sum_{t=0}^\infty \gamma^t r_a^{(t)}\right],
  \label{eqn:pixelrl_return}
\end{equation}
where $\gamma$ is the discount rate, and $\theta$ denotes FCN's parameters.

Policy gradient methods are utilized to maximize $J(\theta)$ by taking steps $\theta \leftarrow \theta + \alpha \nabla_\theta J(\theta)$ iteratively ($\alpha$ is the learning rate).
According to the policy gradient theorem \cite{sutton2018}, the gradient of $J(\theta)$ with respect to $\theta$ equals:
\begin{equation}
  \nabla_\theta J(\theta)
  =\mathbb{E}_{\bpi_\theta} \left[\frac{1}{N} \sum_{a=1}^N \sum_{t=0}^\infty \nabla_\theta \log \pi_a\left(h_a^{(t)}|s_a^{(t)};\theta\right) R_a^{(t)} \right].
  \label{eqn:pixelrl_reinforce}
\end{equation}
The one-sample Monte Carlo estimation of Eq.~\eqref{eqn:pixelrl_reinforce} is the well-known REINFORCE algorithm \cite{williams1992}.
Although it is unbiased, this estimator has a large variance, leading to unstable training.
To stabilize the learning, \cite{furata2020} extended the asynchronous advantage actor-critic (A3C) \cite{mnih2016} for the pixelRL problem.
Specifically, a critic network (a FCN with parameters $\phi$) is introduced to predict the value function $V(\bs^{(t)})=\mathbb{E}_{\bpi_\theta}\left[\mathbf{R}^{(t)}|\bs^{(t)}\right]$, which serves as a \emph{baseline} $b(\bs^{(t)})=V(\bs^{(t)})$ to reduce the variance in the sample estimate for the policy gradient:
\begin{equation}
  R_a^{(t)} = r_a^{(t)} + \gamma V\left(\bs_a^{(t+1)};\phi\right) ,
\end{equation}
\begin{equation}
  A\left(h_a^{(t)},\bs_a^{(t)}\right) = R_a^{(t)} - V\left(\bs_a^{(t)};\phi\right) ,
  \label{eqn:pixelrl_advantage}
\end{equation}
\begin{small}
  \begin{equation}
    \nabla_\theta J(\theta)
    =\mathbb{E}_{\bpi_\theta} \left[\frac{1}{N} \sum_{a=1}^N \sum_{t=0}^\infty \nabla_\theta \log \pi_a(h_a^{(t)}|s_a^{(t)};\theta) A(h_a^{(t)},\bs_a^{(t)}) \right] .
  \end{equation}
\end{small}%
The critic network is dynamically updated to track the current policy network $\bpi_\theta$:
\begin{equation}
  \phi \leftarrow \phi - \alpha \frac{1}{N} \sum_{a=1}^N \nabla_\phi \left( R_a^{(t)}-V\left(\bs_a^{(t)};\phi\right) \right)^2 .
  \label{eqn:pixelrl_update_critic}
\end{equation}

The pixelRL setting is interpretable because the agents' choices in each time step can be observed and analyzed.
For more details of pixelRL, please see \cite{furata2020}.
Next, we shall formally confront the halftoning problem and analyze our DRL-based method's benefits.

\section{Halftoning via DRL}
\label{sec:learn2ht}

\begin{figure*}[t]
  \centering
  \includegraphics[width=0.7\linewidth]{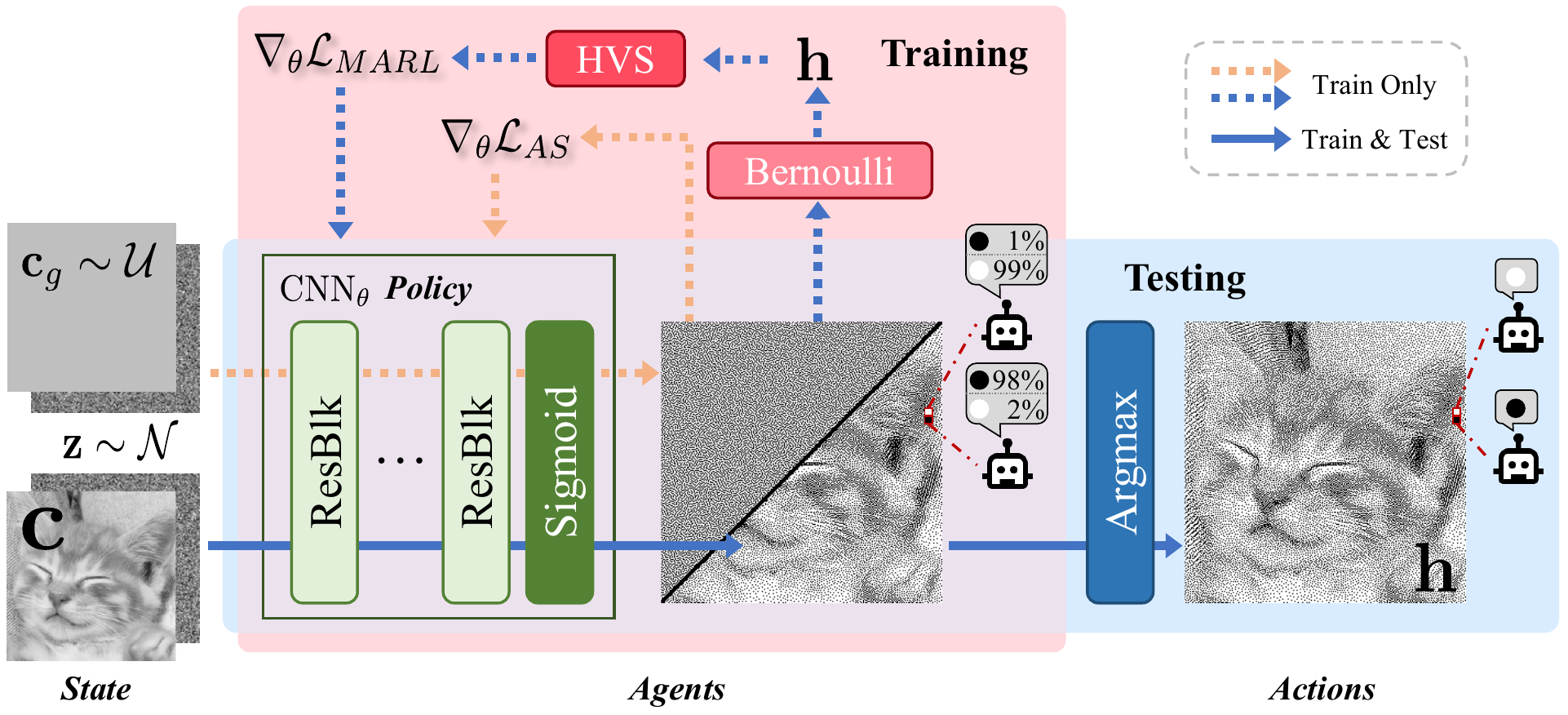}
  \caption{An overview of our proposed deep halftoning framework. The computation of the halftone metric and the analysis of anisotropy only exist in the training phase. The halftone image can be generated by one CNN inference with the trained model.}
  \label{fig:overview}
\end{figure*}

Given a continuous-tone image $\bc \in \mathcal{C}$, a search-based halftoning algorithm \cite{analoui1992,pang2008} generates a halftone image $\bh \in \{0,1\}^N$ by minimizing the designed quantitative visual error $E(\bh,\bc)$, where $\mathcal{C}$ is the contone image dataset, $N$ is the number of pixels, and $E(\cdot,\cdot)$ is a predefined error metric.
Binary pixels $h_a \in \{0,1\}$ compose $\bh$, in which $a \in \{1,\ldots,N\}$ identifies the pixel index.
Instead of performing individual computationally expensive searching for one image at runtime, we \emph{amortize} the cost by training a deep neural network to perform halftoning.
Similar to the pixelRL \cite{furata2020}, we formulate the halftoning problem as a fully cooperative multi-agent reinforcement learning problem.
But here's the key difference: in our formulation, all agents act in just one step, so the halftoning can be done by one NN forward computation.
In other words, the first set of actions made by the agents is the final halftone image, and there is no need for iterative improvements.
This single-step setting also enables a simple yet effective training scheme, which we shall show later in this section.
In the training phase (offline), agents are trained to generate high quality halftones by the RL algorithm.
Then, in the testing phase, the halftone image is generated by one inference, which can be fast if a light-weight model is adopted (see Fig.~\ref{fig:overview}).
Now we cast halftoning into the RL terminology.

\textbf{State.}
The environment state $\mathbf{s}$ is defined as:
\begin{equation}
  \mathbf{s} = \text{Concatenate}(\bc,\bz),
\end{equation}
where $\bz$ is a dynamically sampled white Gaussian noise map enabling a CNN to dither constant grayness images \cite{xia2021}.

\textbf{Agents.}
A virtual agent with a shared policy $\pi$ is constructed at each pixel's position, and there are $N$ agents in total.
Specifically, we utilize a FCN as the policy network like the pixelRL \cite{furata2020}.
After observing the state $\bs_a$ (the receptive field of an output pixel $a$) and communicating with other agents during the CNN forward pass, agent at position $a$ gives its probabilities of actions:
\begin{equation}
  \text{Pr}(h_a|\bc,\bz) = \pi_a(h_a|\bc,\bz;\theta),
\end{equation}
where $\theta$ denotes CNN's parameters.
The joint policy is denoted as $\bpi = \{\pi_1,\ldots,\pi_N\}$.
We think that the noise map $\bz$, which is proposed by \cite{kim2018} and \cite{xia2021}, plays as a disentangled latent variable or an image-agnostic ``plan''.
It indicates the specific halftone image that should be predicted in this multimodal problem, which means the halftone pixels are independent of each other conditioning on $\bc$ and $\bz$:
\begin{equation}
  \bpi(\bh|\bc,\bz;\theta) = \prod_a \pi_a(h_a | \bc,\bz;\theta).
  \label{eqn:independent}
\end{equation}
In other words, actions (pixels) can be chosen simultaneously by agents in both training and testing.
On the contrary, error diffusion and search-based methods process pixels in an autoregressive-like manner \cite{dahl2017}: $\text{Pr}(\bh|\bc) = \prod_{i=1}^N \text{Pr}(h_i|\bh_{<i}, \bc)$, whose parallelism is extremely limited.

\begin{figure}[t]
  \centering
  \includegraphics[width=0.8\linewidth]{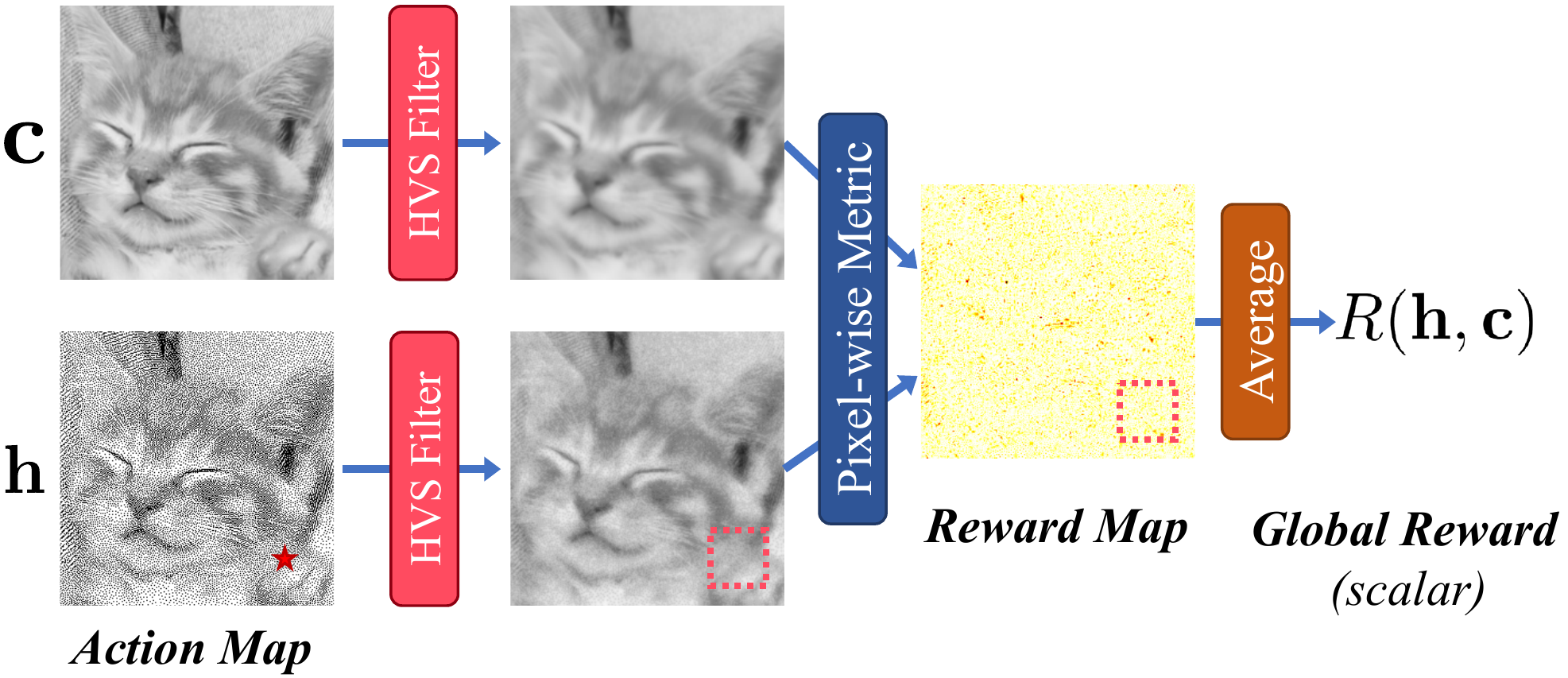}
  \caption{Illustration of halftone assessing steps and the reward map. An agent's action (marked by ``$\color{red}\bigstar$'') can only affect the rewards in the window around it, whose size is the same as the HVS filter.}
  \label{fig:reward_map}
\end{figure}

\textbf{Rewards.}
In halftone assessment, existing metrics (MSE and SSIM) consist of three steps (shown in Fig.~\ref{fig:reward_map}):
(1) Filter $\bh$ and $\bc$ using the HVS model.
(2) Calculate the pixel-wise error map $\be$.
(3) Calculate the scalar metric by averaging the error map.
Following RL terminology, we define the reward, which we would like to maximize, as the negative of the error.
So there is a reward map $\br$, and the \emph{global} reward $R$ equals:
\begin{equation}
  \label{eqn:reward}
  R(\bh,\bc) = \frac{1}{N} \sum_{a=1}^N r_a = -E(\bh,\bc) = \frac{1}{N} \sum_{a=1}^N -e_a .
\end{equation}
Strictly speaking, an agent's action will affect all the reward values in the local window around it (see the red dashed box in Fig.~\ref{fig:reward_map}), whose size is decided by the HVS filter.
However, since HVS model is usually a low-pass filter, one agent should bear more of the responsibility for those reward values that are closer to it.
This is called the multi-agent credit assignment problem in the RL literature \cite{gronauer2022}.
The scalar performance measure that we want to maximize is:
\begin{equation}
  \label{eqn:return}
  J(\theta) = \mathbb{E}_{\bc,\bz}\left[\mathbb{E}_{\bh \sim \bpi(\bh|\bc,\bz;\theta)}\left[R(\bh,\bc)\right]\right].
\end{equation}
Since there is only one decision step in our formulation, Eq.~\eqref{eqn:return} does not involve the integration over time steps, which is different to Eq.~\eqref{eqn:pixelrl_return}.
The specific form of the reward function $R(\cdot,\cdot)$ will be discussed in Section~\ref{sec:metric}.

\textbf{Learning.}
In order to maximize $J(\theta)$ (Eq.~\eqref{eqn:return}), we adopt policy gradient-based methods.
The gradient equals:
\begin{align}
  \nabla_\theta J(\theta)
   & = \nabla_\theta \mathbb{E}_{\bc,\bz}\left[\mathbb{E}_{\bh \sim \bpi(\bh|\bc,\bz;\theta)}\left[R(\bh,\bc)\right]\right] \notag \\
   & = \mathbb{E}_{\bc,\bz}\left[\sum_{\bh} \nabla_\theta \bpi(\bh|\bc,\bz;\theta) R(\bh,\bc)\right] .
  \label{eqn:gradient}
\end{align}
We estimate this expectation by taking $B$ samples (mini-batch) from the dataset $\mathcal{C}$ and the multi-variate Gaussian prior distribution:
\begin{gather}
  \nabla_\theta J(\theta) \approx \frac{1}{B} \sum_{i=1}^B g(\bc^{(i)},\bz^{(i)}), \\
  g(\bc,\bz) = \sum_{\bh} \nabla_\theta \bpi(\bh|\bc,\bz;\theta) R(\bh,\bc),
  \label{eqn:inner_grad}
\end{gather}
where $\bc^{(i)}\sim\mathcal{C}$ and $\bz^{(i)}\sim\mathcal{N}(\bzero,\bI)$.
However, the exact integration of $g(\bc,\bz)$ (Eq.~\eqref{eqn:inner_grad}) is still intractable (with $O(2^N)$ complexity of reward computation).
Therefore, we use a gradient estimator $\hat{g} \approx g$ via Monte Carlo estimation \cite{mohamed2020}.

Next, we will discuss how we design $\hat{g}$.
To keep the notation simple, we leave it implicit that $\pi$ is a function of $\bc$, $\bz$, $\theta$; $R$, $g$ are functions of $\bc$, $\bz$; and the gradient is with respect to $\theta$.

\subsection{REINFORCE with Counterfactual Baseline}
The REINFORCE \cite{williams1992} algorithm used in Eq.~\eqref{eqn:pixelrl_reinforce} is still feasible.
However, here we do not need the policy gradient theorem \cite{sutton2018} due to the single step setting, and the gradient estimator can be directly derived by the log derivative trick:
\begin{align}
  g \notag
   & = \sum_{\bh} \nabla \bpi(\bh) R(\bh) \notag                                                          \\
   & = \sum_{\bh} \bpi(\bh) \nabla \log \bpi(\bh) R(\bh) \notag                                           \\
   & = \mathbb{E}_{\bh\sim\bpi} \left[ \nabla \left( \log \prod_a \pi_a(h_a) \right)R(\bh) \right] \notag \\
   & = \mathbb{E}_{\bh\sim\bpi} \left[ \sum_a \nabla \log \pi_a(h_a) \left(R(\bh) - b_a\right)\right] .
  \label{eqn:reinforce}
\end{align}
In pixelRL \cite{furata2020}, a critic network \cite{mnih2016} is adopted to predict the expected accumulated reward (value function) as the baseline $b_a$, and this additional model is simultaneously updated within the training procedure (see Eq.~\eqref{eqn:pixelrl_advantage} and \eqref{eqn:pixelrl_update_critic}).
But here we find that it is possible to directly calculate a more precise baseline.
Specifically, we introduce the counterfactual multi-agent (COMA) baseline \cite{foerster2018}:
\begin{equation}
  \label{equ:baseline}
  b_a=b_a(\bh_{-a}) = \sum_{h_a'} \pi_a(h_a') R(\{h_a',\bh_{-a}\}),
\end{equation}
where $\bh_{-a} = \{h_1,\ldots,h_{a-1},h_{a+1},\ldots,h_{N}\}$.
Intuitively, this baseline $b_a(\bh_{-a})$ indicates agent $a$'s average performance by marginalizing its action $h_a'$, while keeping other agents' actions $\bh_{-a}$ fixed.
It is computationally feasible in the halftoning problem, as the dimension of a single agent's action space is only $2$ (black and white) and the reward function has locality (see the window in Fig.~\ref{fig:reward_map}).
More importantly, it elegantly tackles the credit assignment problem aforementioned by counterfactual thinking: ``what if an agent had chosen the other action''.
Combining Eq.~\eqref{eqn:reinforce} and \eqref{equ:baseline}, the one-sample ($\bh\sim\bpi$) COMA gradient estimator is:
\begin{equation}
  \label{eqn:coma}
  \hat{g}_{\text{COMA}} = \sum_a \nabla \log \pi_a(h_a) \left(R(\bh) - b_a(\bh_{-a}) \right) .
\end{equation}

\begin{figure}[t]
  \centering
  \includegraphics[width=\linewidth]{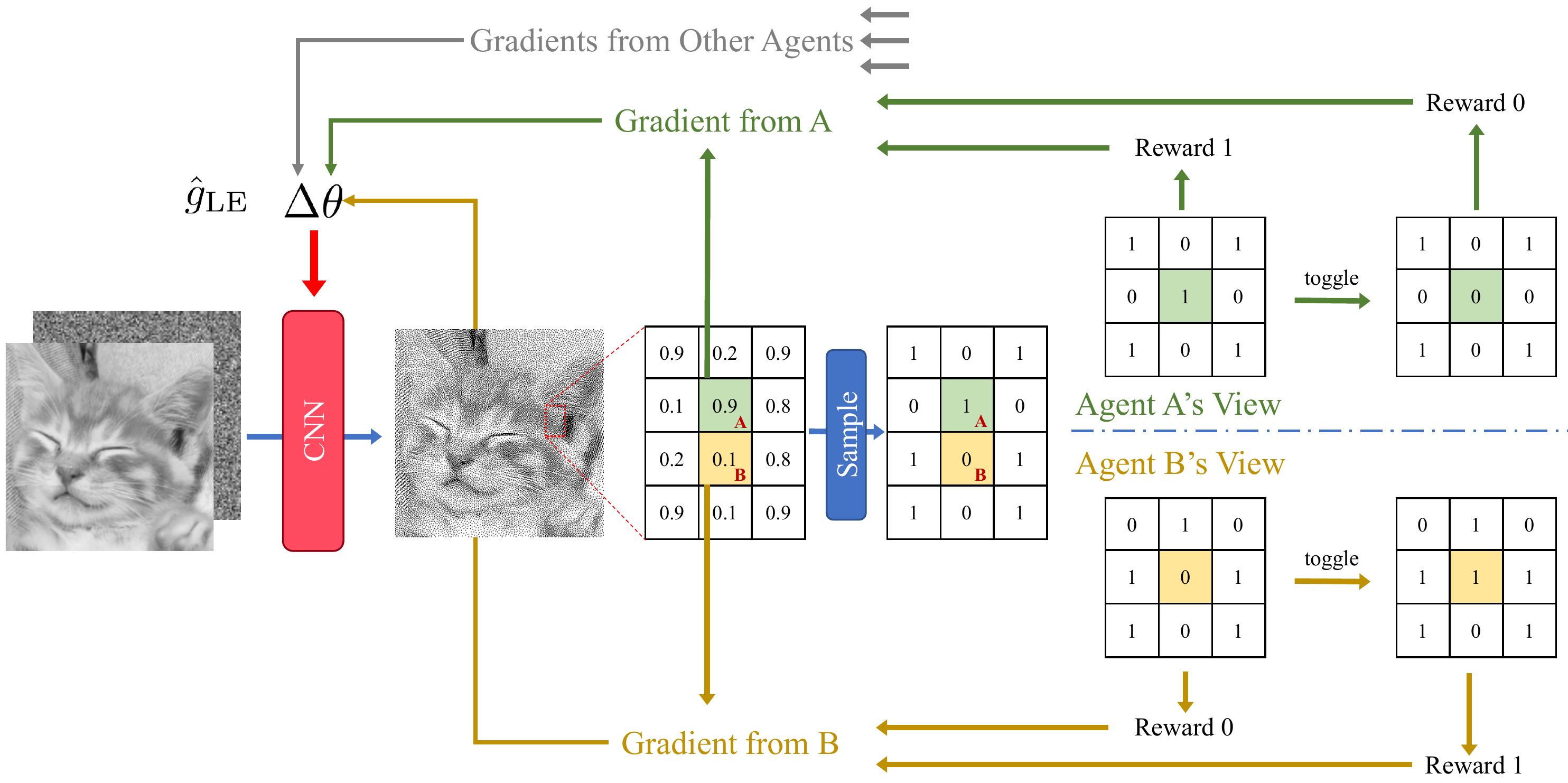}
  \caption{Illustration of $\hat{g}_{\text{LE}}$ (Eq.~\eqref{eqn:le}).}
  \label{fig:grad_illustration}
\end{figure}

\subsection{Local Expectation Gradient Estimator}
The gradient estimator $\hat{g}_{\text{COMA}}$ (Eq.~\eqref{eqn:coma}) is a feasible solution with one sample.
However, we notice that the COMA baseline (Eq.~\eqref{equ:baseline}) actually has explored $N$ extra non-duplicate data points.
It may reduce more variance, if we directly utilize these data points to estimate Eq.~\eqref{eqn:inner_grad}, rather than regard them as a part of the baseline.
Following this idea, we derive a new estimator with $N+1$ evaluation budgets of $R(\cdot)$:
\begin{align}
  \label{eqn:halftoners}
   & g = \sum_\bh \nabla\bpi(\bh) R(\bh) \notag                                                                                \\
   & = \sum_{\bh}\sum_a \nabla\pi_a(h_a)\bpi_{-a}(\bh_{-a})R(\{h_a,\bh_{-a}\}) \notag                                          \\
   & = \sum_a \sum_{h_a}\sum_{\bh_{-a}} \nabla\pi_a(h_a)\bpi_{-a}(\bh_{-a})R(\{h_a,\bh_{-a}\})\sum_{h_a'}\pi_a(h_a') \notag    \\
   & = \sum_a \sum_{h_a'} \sum_{\bh_{-a}} \pi_a(h_a')\bpi_{-a}(\bh_{-a}) \sum_{h_a} \nabla\pi_a(h_a)R(\{h_a,\bh_{-a}\}) \notag \\
   & = \sum_{\bh'} \bpi(\bh') \sum_a \sum_{h_a} \nabla\pi_a(h_a)R(\{h_a,\bh_{-a}'\}) \notag                                    \\
   & = \mathbb{E}_{\bh \sim \bpi}\left[\sum_a \sum_{h_a'} \nabla\pi_a(h_a')R(\{h_a',\bh_{-a}\})\right] .
\end{align}
This is similar to \cite{titsias2015}, which was proposed to estimate the gradient of the evidence lower bound in variational inference problems.
Despite there being some differences (\eg, our agents are independent conditioning on the shared $\theta$, while the $\bh$ was represented as a directed graph in their work), we adopt their name: local expectation gradient, since the spirit under the hood is similar: calculating a local exact expectation while using a single sample from the other variables.
The corresponding one-sample estimator is:
\begin{equation}
  \label{eqn:le}
  \hat{g}_{\text{LE}} = \sum_a \sum_{h_a'} \nabla\pi_a(h_a')R(\{h_a',\bh_{-a}\}) ,
\end{equation}
where $\bh\sim\bpi$.
While the computational complexity is the same: $O(N+1)$, we empirically find that $\hat{g}_{\text{LE}}$ gets better optimization results than $\hat{g}_{\text{COMA}}$ (see Section~\ref{sec:exps}(D)).

Besides, Eq.~\eqref{eqn:le} has an intuitive explanation.
It is similar to performing $N$ concurrent ``toggle'' operations from the DBS algorithm \cite{analoui1992} on the sampled halftone image.
But instead of updating the halftone in place, we turn the reward differences into gradients to reinforce the NN (illustrated in Fig.~\ref{fig:grad_illustration}).
The complete formula is given here, as our final proposal:

\begin{small}
  \begin{equation}
    \label{eqn:marl}
    \nabla_\theta \mathcal{L}_{MARL} = -\mathbb{E}_{\bc,\bz,\bh\sim\bpi}\left[\sum_a \sum_{h_a'} \nabla\pi_a(h_a')R(\{h_a',\bh_{-a}\},\bc)\right].
  \end{equation}
\end{small}%
We emphasize that no additional NNs, reference halftone datasets, hyper-parameters, or heuristic binarization rules are needed in $\nabla_\theta \mathcal{L}_{MARL}$.

\subsection{Efficient Calculation of The Policy Gradient}
One may argue that the $N+1$ times of $R(\cdot)$ in Eq.~\eqref{eqn:marl} are too expensive.
Here we point out that it can be efficient when considering the characteristics of halftone's metrics (see Fig.~\ref{fig:reward_map}).
First, all agents can reuse the same HVS filtered map.
Second, an agent only needs to be responsible for the local window around it on the reward map.
Third, instead of calculating $R(\cdot)$ from scratch, the \emph{opposite} action's reward window equals the \emph{current} action's reward window plus/minus the HVS filter, from each agent's perspective.
In a word, the computation of Eq.~\eqref{eqn:marl} is doable and parallelizable, and all the training cost only appears in the offline phase.

\begin{figure*}[t]
  \begin{minipage}[b]{0.015\linewidth}
    \centering
    \footnotesize{\rotatebox{90}{\hspace{5.3em}Anisotropy\hspace{6em}RAPSD\hspace{2em}MSE}}
  \end{minipage}
  \hfill
  \begin{minipage}[b]{0.98\linewidth}
    \centering
    \begin{minipage}[b]{0.24\linewidth}
      \centering
      \includegraphics[width=\linewidth]{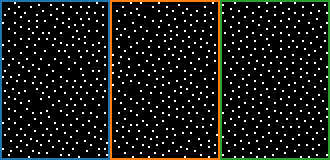}
      \vfill
      \small{5.69e-4 / 5.76e-4 / 4.87e-4}
    \end{minipage}
    \hfill
    \begin{minipage}[b]{0.24\linewidth}
      \centering
      \includegraphics[width=\linewidth]{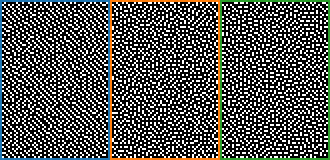}
      \vfill
      \small{5.28e-4 / 5.92e-4 / 4.20e-4}
    \end{minipage}
    \hfill
    \begin{minipage}[b]{0.24\linewidth}
      \centering
      \includegraphics[width=\linewidth]{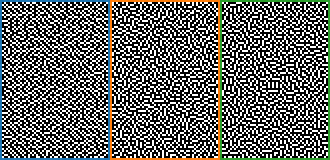}
      \vfill
      \small{7.04e-4 / 6.73e-4 / 4.68e-4}
    \end{minipage}
    \hfill
    \begin{minipage}[b]{0.24\linewidth}
      \centering
      \includegraphics[width=\linewidth]{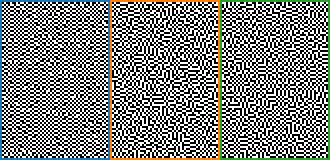}
      \vfill
      \small{3.45e-4 / 5.40e-4 / 3.41e-4}
    \end{minipage}
    \vfill
    \medskip
    \begin{minipage}[b]{0.24\linewidth}
      \centering
      \includegraphics[width=\linewidth]{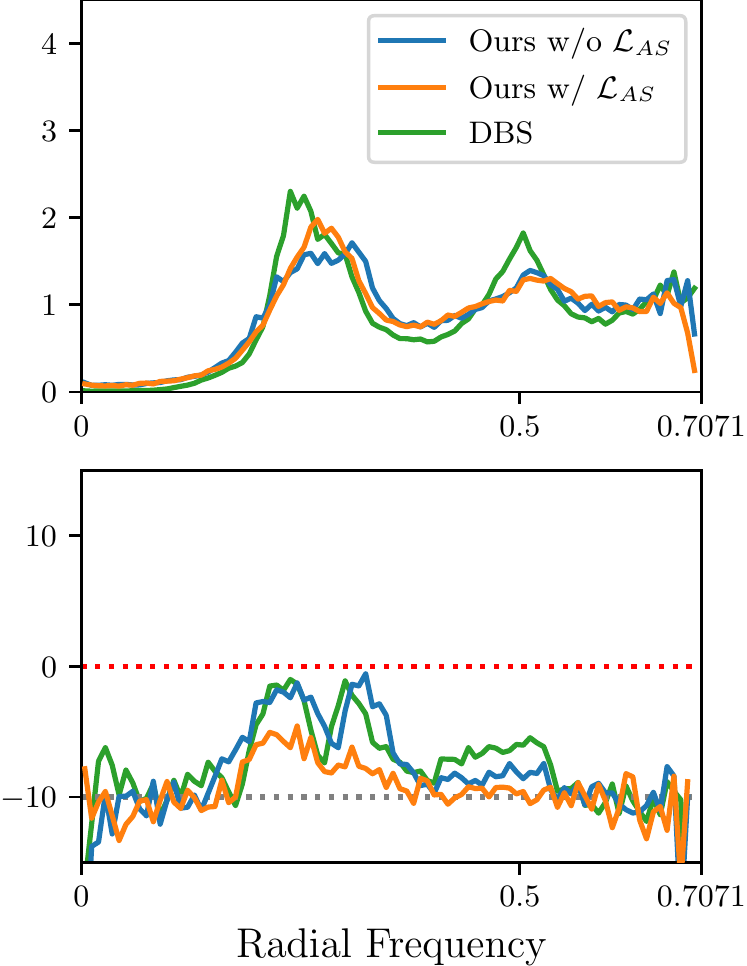}
      \vfill
      \small{(a) g=15/255}
    \end{minipage}
    \hfill
    \begin{minipage}[b]{0.24\linewidth}
      \centering
      \includegraphics[width=\linewidth]{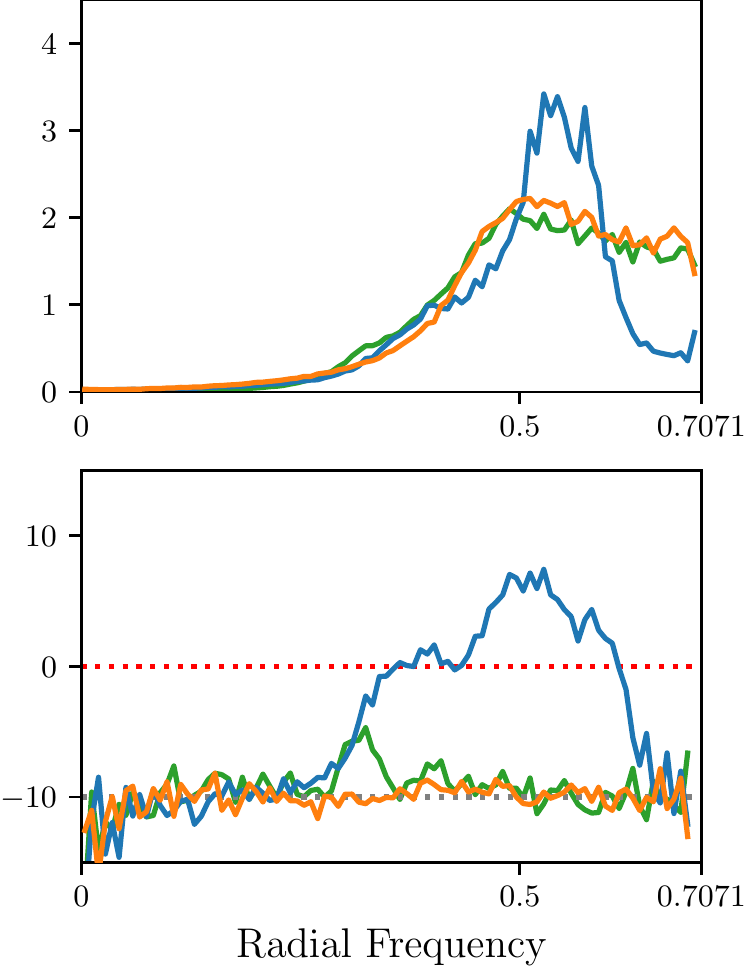}
      \vfill
      \small{(b) g=80/255}
    \end{minipage}
    \hfill
    \begin{minipage}[b]{0.24\linewidth}
      \centering
      \includegraphics[width=\linewidth]{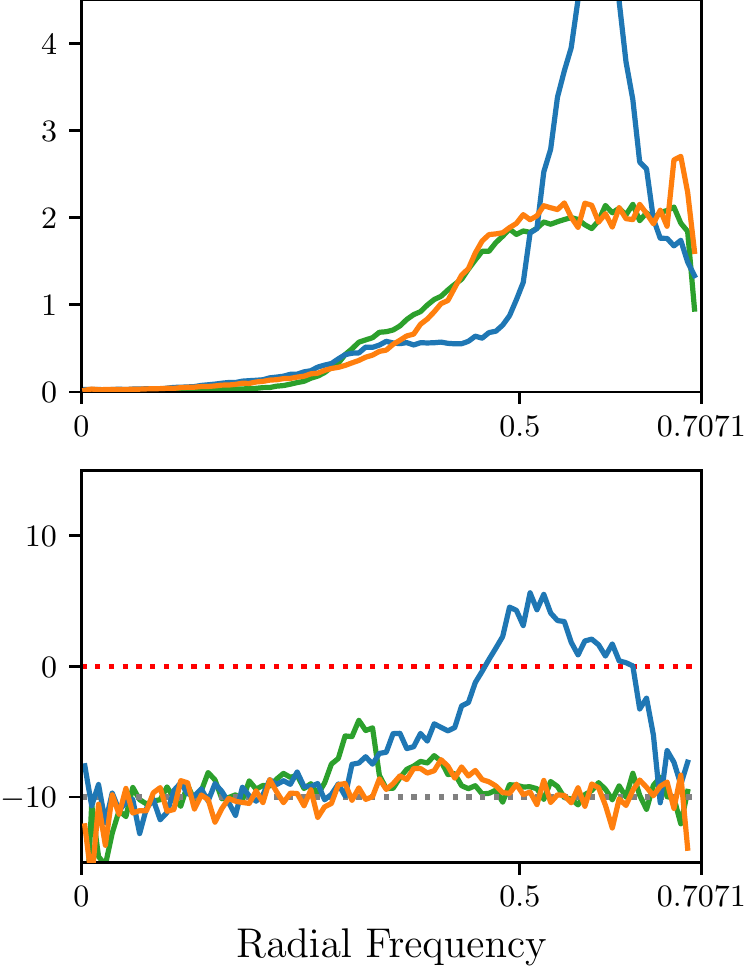}
      \vfill
      \small{(c) g=100/255}
    \end{minipage}
    \hfill
    \begin{minipage}[b]{0.24\linewidth}
      \centering
      \includegraphics[width=\linewidth]{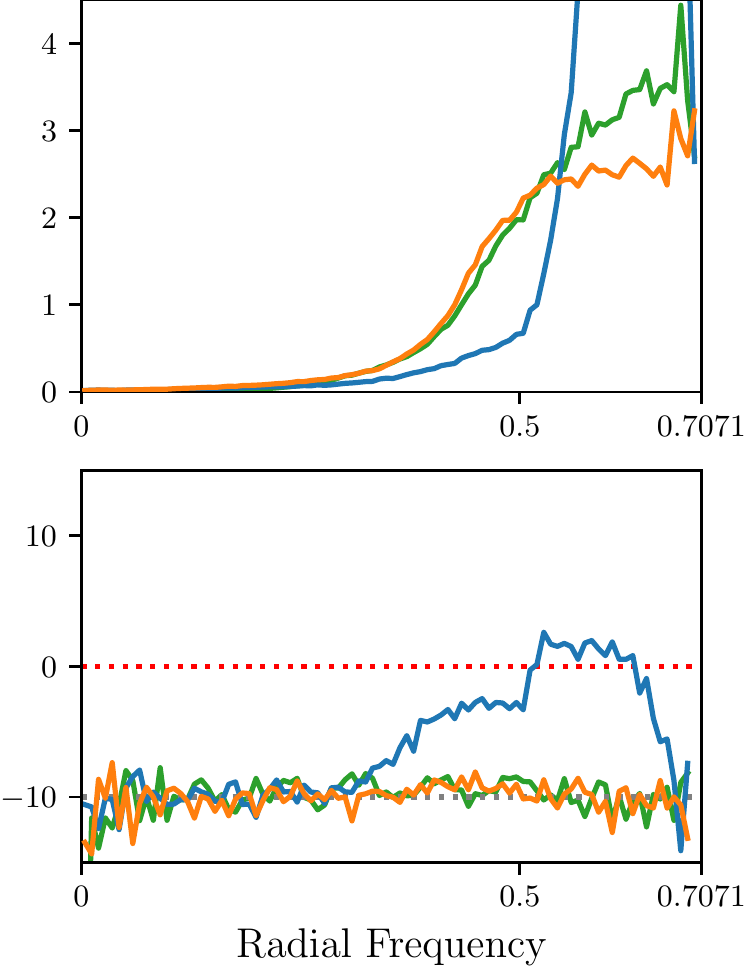}
      \vfill
      \small{(d) g=127/255}
    \end{minipage}
  \end{minipage}
  \caption{Spectral analyses on our halftone results generated without $\mathcal{L}_{AS}$ and with $\mathcal{L}_{AS}$. The corresponding MSE scores measured on the test set are 5.01e-4 and 5.51e-4, respectively. Results from the DBS algorithm \cite{analoui1992} are also included here for reference. Halftone samples in each column: (left) ours w/o $\mathcal{L}_{AS}$, (middle) ours w/ $\mathcal{L}_{AS}$, (right) DBS.}
  \label{fig:bluenoise}
\end{figure*}

\section{Anisotropy Suppressing Loss Function}
\label{sec:learn2blue}
Although our proposed DRL-based model enables training CNNs to output discrete halftone images, the desirable blue-noise property \cite{ulichney1988} has not been explicitly guaranteed.
Following classic search-based methods \cite{analoui1992}, we tested the MSE metric (between the HVS-filtered halftone and HVS-filtered contone) as the reward function in the proposed DRL framework.
However, we find that a lower MSE score does not necessarily mean the better blue-noise quality (see Fig.~\ref{fig:bluenoise}).
Even worse, the parallelism of CNN makes it easy to create globally consistent stripe and checkerboard-like artifacts.

These phenomena were also reported by Xia \etal \cite{xia2021}.
In response, they proposed to penalize the low frequency components, separated by the discrete cosine transform, on the dithered constant grayness images.
However, we observed that the low-frequency components have been minimized well by the proposed policy gradient.
The more urgent defect here is the excessive anisotropy, which was neither explicitly considered by MSE nor SSIM.
According to \cite{lau2008}, the anisotropy of a constant grayness image's blue-noise halftone should be minimized to $-10\text{dB}$ at all frequencies.
Thus we intend to explicitly suppress the anisotropy.

Given the definition of the power spectral estimate $\hat{P}(f)$ with one periodogram and the radially averaged power spectrum density (RAPSD) $P(f_\rho)$ \cite{ulichney1988}:
\begin{gather}
  \label{eqn:psd}
  \hat{P}(f) \approx \frac{1}{N} \lvert \text{DFT}(\bh) \rvert^2 \\
  \label{eqn:rapsd}
  P(f_\rho) = \frac{1}{n(r(f_\rho))} \sum_{f \in r(f_\rho)} \hat{P}(f),
\end{gather}
where $n(r(f_\rho))$ is the number of discrete frequency samples in an annular ring with width $\Delta_\rho = 1$ around radial frequency $f_\rho$, the anisotropy is defined as:
\begin{equation}
  A(f_\rho) = \frac{1}{n(r(f_\rho))-1} \sum_{f \in r(f_\rho)} \frac{\left(\hat{P}(f)-P(f_\rho)\right)^2}{P^2(f_\rho)}.
  \label{eqn:anis}
\end{equation}
To achieve the blue-noise property, we propose the anisotropy suppressing loss function:
\begin{equation}
  \label{eqn:as}
  \mathcal{L}_{AS} = \mathbb{E}_{\bc,\bz} \left[\sum_{f_\rho}\sum_{f\in r(f_\rho)}\left(\hat{P}_\theta(f) - P_\theta(f_\rho)\right)^2\right],
\end{equation}
which minimizes the numerator of Eq.~\eqref{eqn:anis}.
Note that the calculation of $\hat{P}_\theta(f)$ involves the $\text{Thresholding}(\cdot)$ operation, which also leads to the gradient vanishing problem.
Naturally, one may want to treat Eq.~\eqref{eqn:as} as part of the reward function (Eq.~\eqref{eqn:reward}) in the RL framework.
But the fatal drawback here is: a single pixel's action switching will lead to changes on the whole spectrum.
So the naïve implementation of $\hat{g}_{LE}$ optimizing $\mathcal{L}_{AS}$ will bring $N+1$ times of anisotropy calculation in every training iteration, which is unacceptable in practice.

\begin{figure}[t]
  \centering
  \begin{minipage}{0.3\linewidth}
    \centering
    \includegraphics[width=\linewidth]{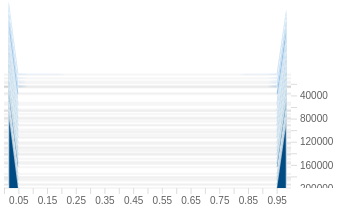}
    \vfill
    \small{(a)}
    \vfill
    \includegraphics[width=\linewidth]{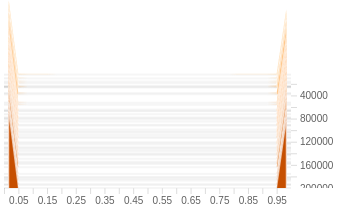}
    \vfill
    \small{(b)}
  \end{minipage}
  \begin{minipage}{0.5\linewidth}
    \centering
    \includegraphics[width=\linewidth]{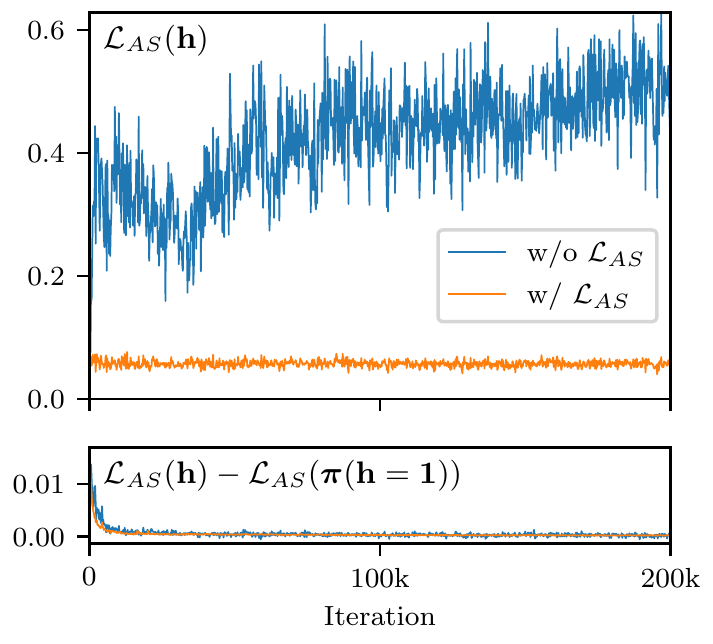}
    \vfill
    \small{(c)}
  \end{minipage}
  \caption{The histograms of CNN's outputs (analyzed on the Lenna test image) during the training (a) w/o $\mathcal{L}_{AS}$ and (b) w/ $\mathcal{L}_{AS}$. (c) Curves of $\mathcal{L}_{AS}$.}
  \label{fig:anis_proxy}
\end{figure}

To address this problem, we make the following key observation: most of the output values of our network are clustered around the two endpoints in the whole training process (shown in Fig.~\ref{fig:anis_proxy}(a)).
Therefore the probability map of action ``white'' should be similar to the real halftone image (after thresholding).
According to this observation, we take the differentiable probabilities $\bpi(\bh=\bone)$ as a proxy for $\bh$ to estimate the power spectrum in Eq.~\eqref{eqn:psd}.
As the spectral analysis only makes sense for constant grayscale images' halftones, we optimize this loss function on an extra mini-batch of uniformly sampled $\bc_g \sim \mathcal{U}(\bzero,\bone)$ following \cite{xia2021}.
Our experimental results show the artifacts vanish as a result of $\mathcal{L}_{AS}$. The clustered distribution still holds after adding the new loss function (see Fig.~\ref{fig:bluenoise}(b)).

Now we are able to train a deep blue-noise halftoning model via gradient descending (see Algorithm~\ref{alg:alg1} and Fig.~\ref{fig:overview}):
\begin{equation}
  \nabla \mathcal{L}_{total} = \nabla \mathcal{L}_{MARL} + w_a \nabla \mathcal{L}_{AS},
\end{equation}
where $w_a$ is a hyper-parameter.
The symbols defined in Section~\ref{sec:learn2ht} and \ref{sec:learn2blue} are summarized in Table~\ref{tab:notations}.
In the next section, we are going to discuss some drawbacks of existing halftone metrics.

\begin{algorithm}[h!]
  \caption{Halftoning via Deep Reinforcement Learning.}
  \begin{algorithmic}
    \Procedure{\textsc{Train}}{$\mathcal{C}$}
    \State Init $\theta$
    \Repeat
    \LComment{Optimize halftone metrics}
    \State Sample $\bc \sim \mathcal{C}$, $\bz \sim \mathcal{N}(\bzero,\bI)$
    \State $\bh \sim \bpi(\bh|\bc,\bz;\theta)$
    \State \small{$\mathcal{L}_{MARL} = -\sum_a \sum_{h_a'}R(\{h_a',\bh_{-a}\},\bc)\pi_a(h_a'|\bc,\bz;\theta)$}
    \LComment{Suppress anisotropy}
    \State Sample $\bc_g \sim \mathcal{U}(\bzero,\bone)$, $\bz_g \sim \mathcal{N}(\bzero,\bI)$
    \State $\hat{P}_\theta(f)=\frac{1}{N}|\text{DFT}(\bpi(\bh=\bone|\bc_g,\bz_g;\theta))|^2$
    \State $P_\theta(f_\rho)=\frac{1}{n(r(f_\rho))} \sum_{f \in r(f_\rho)} \hat{P}_\theta(f)$
    \State \small{$\mathcal{L}_{AS} = \sum_{f_\rho}\sum_{f\in r(f_\rho)}\left(\hat{P}_\theta(f)-P_\theta(f_\rho)\right)^2$}
    \LComment{Update weights}
    \State $\theta \leftarrow \theta - \alpha (\nabla_\theta \mathcal{L}_{MARL} + w_a \nabla_\theta \mathcal{L}_{AS})$
    \Until{Convergence}
    \State \Return $\theta$
    \EndProcedure
    \Procedure{\textsc{Test}}{$\bc,\theta$}
    \State Sample $\bz \sim \mathcal{N}(\bzero,\bI)$
    \State $\bh=\text{Thresholding}(\text{CNN}_\theta(\bc,\bz), 0.5)$
    \State \Return $\bh$
    \EndProcedure
  \end{algorithmic}
  \label{alg:alg1}
\end{algorithm}

\begin{table}[h!]
  \centering
  \caption{Table of Notations}
  \label{tab:notations}
  \begin{tabular}{c|c}
    \toprule
    Symbol                     & Meaning                                              \\
    \midrule
    $\text{Pr}(\cdot)$         & Probability                                          \\
    $\bh$                      & Halftone image/actions                               \\
    $\bc$                      & Continuous-tone image                                \\
    $\mathcal{C}$              & Continuous-tone dataset                              \\
    $\mathcal{N}$              & Gaussian distribution                                \\
    $\mathcal{U}$              & Uniform distribution                                 \\
    $\bz$                      & Noise map                                            \\
    $\bs$                      & State                                                \\
    $a$                        & Index of the $a^\text{th}$ agent/pixel               \\
    $N$                        & Number of agents/pixels                              \\
    $\theta$                   & CNN parameters                                       \\
    $\bpi$                     & Policy                                               \\
    $\bpi(\bh|\bc,\bz;\theta)$ & Probability of $\bh$ given $\bc$, $\bz$ and $\theta$ \\
    $E(\cdot,\cdot)$           & Error function                                       \\
    $R(\cdot,\cdot)$           & Reward function                                      \\
    $J(\theta)$                & Performance measure for the policy with $\theta$     \\
    $b_a$                      & Agent $a$'s baseline function                        \\
    $\mathcal{L}$              & Loss function                                        \\
    $f$                        & Frequency                                            \\
    $\hat{P}(f)$               & Power spectral estimate                              \\
    $f_\rho$                   & Radial frequency                                     \\
    $r(f_\rho)$                & Annular rings with center radius $f_\rho$            \\
    $n(r(f_\rho))$             & Number of frequency samples in $r(f_\rho)$           \\
    $P(f_\rho)$                & Radially averaged power spectrum density of $f_\rho$ \\
    $A(f_\rho)$                & Anisotropy of $f_\rho$                               \\
    \bottomrule
  \end{tabular}
\end{table}

\begin{figure*}[t]
  \centering
  \begin{minipage}{0.94\linewidth}
    \centering
    \subfloat[]{\includegraphics[width=0.24\linewidth]{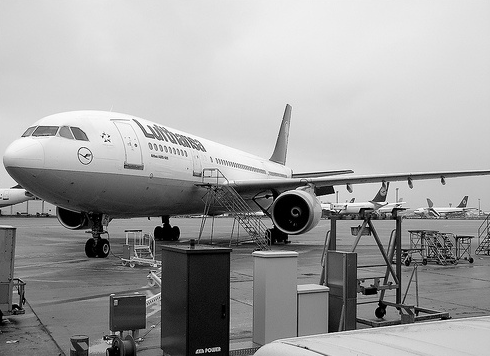}}
    \hfill
    \subfloat[]{\includegraphics[width=0.24\linewidth]{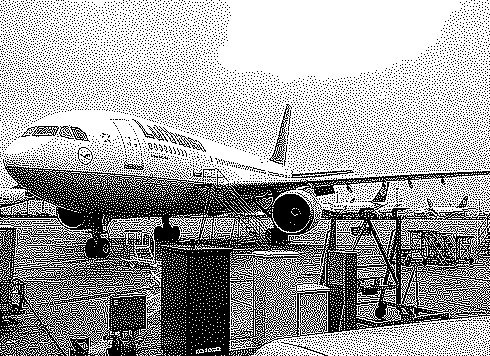}}
    \hfill
    \subfloat[]{\includegraphics[width=0.24\linewidth]{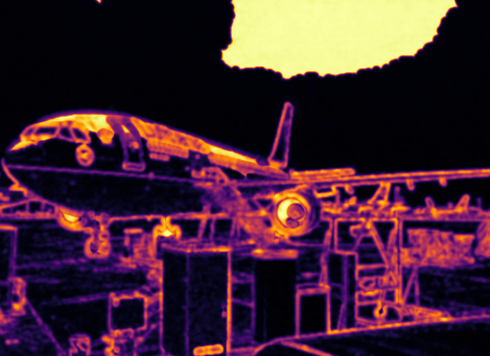}}
    \hfill
    \subfloat[]{\includegraphics[width=0.24\linewidth]{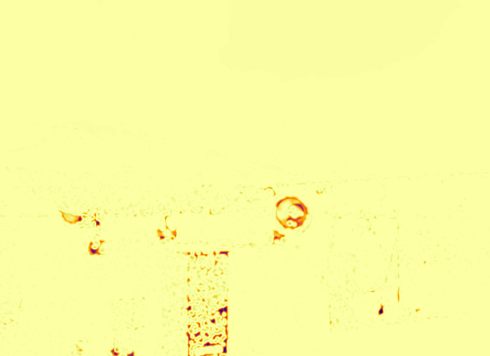}}
    \vfill
    \vspace{-3mm}
    \subfloat[]{\includegraphics[width=0.24\linewidth]{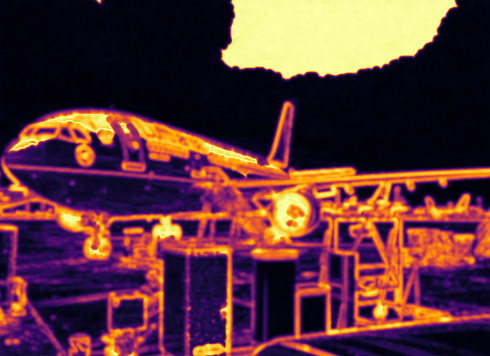}}
    \hfill
    \subfloat[]{\includegraphics[width=0.24\linewidth]{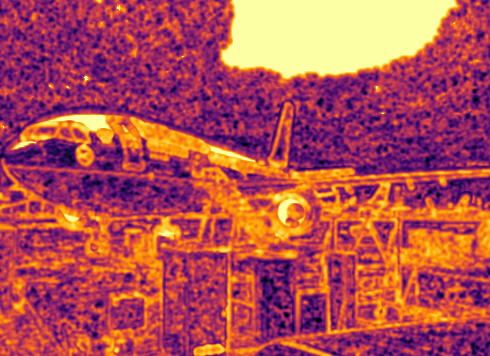}}
    \hfill
    \subfloat[]{\includegraphics[width=0.24\linewidth]{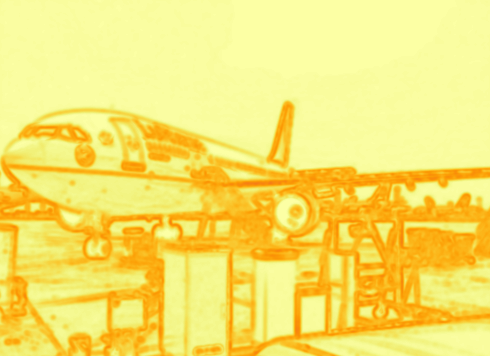}}
    \hfill
    \subfloat[]{\includegraphics[width=0.24\linewidth]{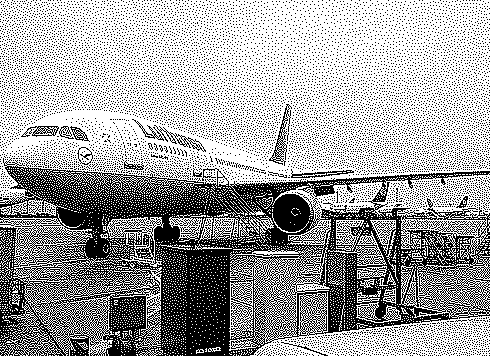}}
  \end{minipage}
  \hfill
  \begin{minipage}{0.05\linewidth}
    \centering
    \includegraphics[height=6cm]{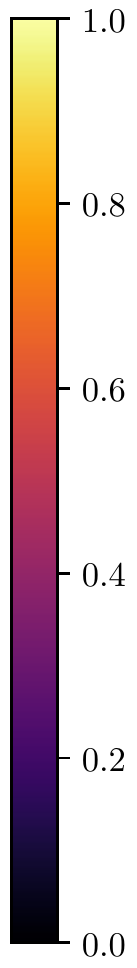}
  \end{minipage}
  \caption{(a) Continuous-tone image.
    (b) Halftone image optimizing MSE \& SSIM.
    (c) SSIM index map.
    (d) Luminance comparison map.
    (e) Contrast comparison map.
    (f) Structure comparison map.
    (g) CSSIM index map.
    (h) Halftone image optimizing MSE \& CSSIM.
  }
  \label{fig:ssim_analysis}
\end{figure*}

\section{Contrast-weighted SSIM for Halftoning}
\label{sec:metric}
To preserve the characteristic look of textured regions in halftone images, Pang \etal \cite{pang2008} proposed structure-aware halftoning that takes the structural similarity index measure (SSIM) \cite{wang2004} into consideration.
However, we find that the optimizing of the original SSIM metric will cause holes in the smooth area whose values are near 0 or 1 (show in Fig.~\ref{fig:ssim_analysis}~(a)(b)).
On the SSIM index map (Fig.~\ref{fig:ssim_analysis}(c)), the hole area is assigned with higher scores.
By contrast, the area to the left of the hole is given a lower score, which is problematic.
To diagnose this problem, we break SSIM into components:
\begin{equation}
  \text{SSIM}(\bh, \bc) = l(\bh, \bc) \cdot c(\bh, \bc) \cdot s(\bh, \bc),
\end{equation}
where $l$, $c$ and $s$ denote luminance, contrast and structure comparison maps (see Fig.~\ref{fig:ssim_analysis}(d)(e)(f)), respectively.
It shows that in the flat area, the contrast and structure components of SSIM actually prevent agents from generating minority dots.

As suggested in \cite{wang2004}, it is possible to compute a weighted average of the SSIM index map according to the specific application.
Recall that the goal of optimizing SSIM in halftoning is to preserve structural details, which should only be performed in the area with abundant textural variation.
Motivated by it, we propose to revise the metric by weighting the original SSIM with the continuous-tone image's contrast map $\boldsymbol{\sigma}_{\bc}$, whose value at position $a$ is:
\begin{equation}
  \sigma_a = k \left(\sum_{i\in \text{window}(a)} w_i (c_i-\mu_a)^2 \right)^{\frac{1}{2}},
\end{equation}
where $k=2$ is a normalizing factor, $w_i$ is the $i$th weight of $11 \times 11$ Gaussian kernel with standard deviation of $1.5$, and $\mu_a$ is the local luminance \cite{wang2004}.
So the contrast-weighted SSIM (CSSIM) metric is:
\begin{equation}
  \text{CSSIM}(\bh,\bc) = \boldsymbol{\sigma}_{\bc} \cdot \text{SSIM}(\bh,\bc)
  + (\bone - \boldsymbol{\sigma}_{\bc}) \cdot \bone.
  \label{eqn:cssim}
\end{equation}
We test this new metric on Fig.~\ref{fig:ssim_analysis}(b), and it successfully assigns reasonable structural scores to halftones (see Fig.~\ref{fig:ssim_analysis}(g)).

Finally, we define the reward function $R(\bh,\bc)$ used in $\mathcal{L}_{MARL}$, which considers both the tone similarity and the structure similarity:
\begin{equation}
  \label{eqn:final_reward}
  R(\bh,\bc) = -\text{MSE}(\text{HVS}(\bh),\text{HVS}(\bc)) + w_s \text{CSSIM}(\bh,\bc),
\end{equation}
where $\text{HVS}(\cdot)$ denotes the low-pass filtering by a HVS model and $w_s$ is a hyper-parameter.
Note that our method is not tied to a specific HVS model.
One can adopt any HVS model according to practical requirements, so long as it has a limited filter size.
In this paper, without loss of generality, we use Näsänen's HVS model \cite{nasanen1984} for demonstration since it was recommended in \cite{kim2002}.
We use the definition and parameters listed in \cite{kim2002}'s Table~I.
Besides, the filter size is set to 11x11.
With regard to the scale parameter, which serves as a free parameter \cite{kim2002}, we choose $S=2000$, as it is similar to the situation when looking at a 24-inch 1080p monitor.
Fig.~\ref{fig:ssim_analysis}(h) shows the result generated by our model optimizing the expectation of Eq.~\eqref{eqn:final_reward}.
One can see that the hole phenomenon has disappeared, and the image presents clear structural details.

\section{Experiments}
\label{sec:exps}
In this section, we are going to show details of our method, compare our work with prior halftoning approaches, analyze the components, and discuss the extensibility.

\subsection{Experiment Settings}
\textbf{Prior Methods.}
We have selected nine typical halftoning methods from different categories for comparison:
\begin{itemize}
  \item Ordered dithering.
        Void-and-cluster (\textbf{VAC}) \cite{ulichney1993}, dither array size = 64x64.
  \item Error diffusion.
        Ostromoukhov's method (\textbf{OVED}) \cite{ostromoukhov2001};
        Structure-aware error diffusion (\textbf{SAED}) \cite{chang2009};
        Tone-dependent error diffusion based on an updated blue-noise model ($\textbf{TDED}_\textbf{BS}$) \cite{fung2016};
        Simple gradient-based error diffusion (\textbf{SGED})\cite{hu2016}.
  \item Search-based methods.
        Direct binary search (\textbf{DBS}) \cite{analoui1992};
        Structure-aware halftoning (\textbf{SAH}) \cite{pang2008}.
        For a fair comparison, we use the same optimization objective as in our method (Eq.~\eqref{eqn:final_reward}. 11x11 Näsänen HVS filter with $S=2000$) when implementing their methods (DBS: MSE; SAH: MSE and CSSIM, $w_s=0.06$).
  \item Deep halftoning.
        To our knowledge, Choi and Allebach's work \cite{choi2022} is the only published paper currently focusing on generating aperiodic dispersed-dot halftones with deep models.
        We implement the conditional GAN (\textbf{cGAN}) for comparison since the autoregressive model is, according to their report, very slow.
        Except for the dataset, all training details strictly follow the settings in \cite{choi2022}.
  \item In addition, we take the reversible halftoning method (\textbf{RVH}) \cite{xia2021} for comparison, as it can generate blue-noise halftones.
        Specifically, we reimplement it by removing the extra reconstruction task that may damage halftones' quality (only the warm-up stage in their paper).
        The experimental results of the full version (with reconstruction) are also presented for reference.
        Except for the warm-up-only version (trained for 45 epochs for convergence), all training details exactly follow their released code.
\end{itemize}

\begin{figure*}[t]
  \centering
  \captionsetup[subfloat]{labelsep=none,format=plain,labelformat=empty}
  \subfloat[\scriptsize{CT}]{\includegraphics[width=0.24\linewidth]{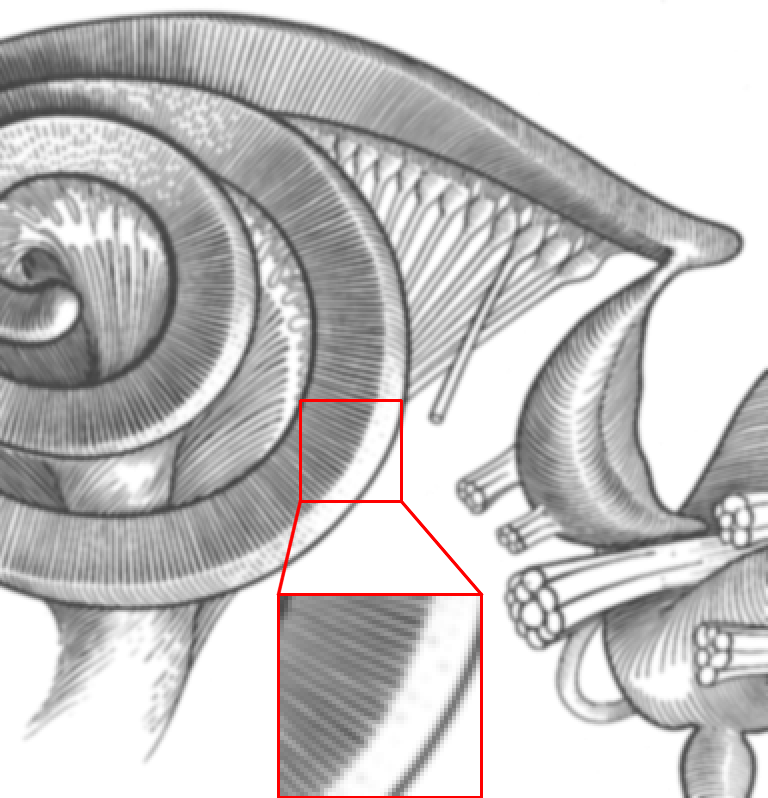}}
  \hspace{1mm}
  \subfloat[\scriptsize{VAC \cite{ulichney1993}}]{\includegraphics[width=0.24\linewidth]{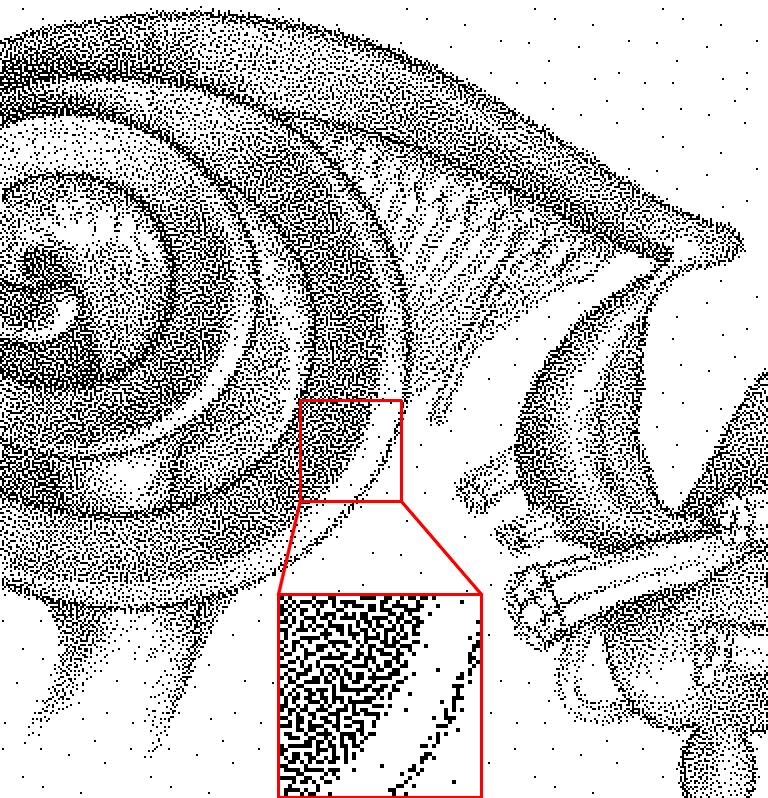}}
  \hspace{1mm}
  \subfloat[\scriptsize{OVED \cite{ostromoukhov2001}}]{\includegraphics[width=0.24\linewidth]{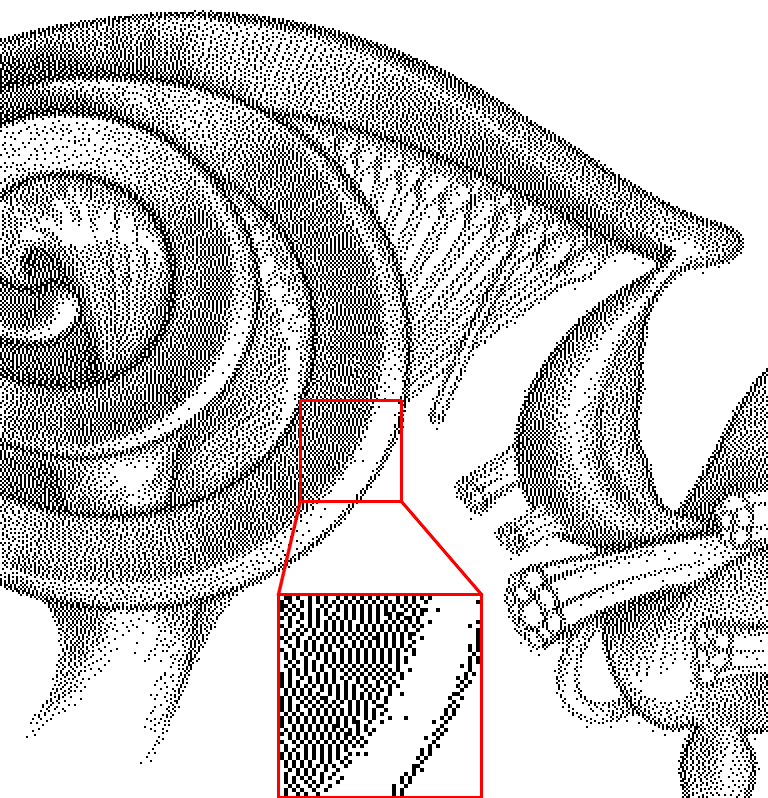}}
  \hspace{1mm}
  \subfloat[\scriptsize{SAED \cite{chang2009}}]{\includegraphics[width=0.24\linewidth]{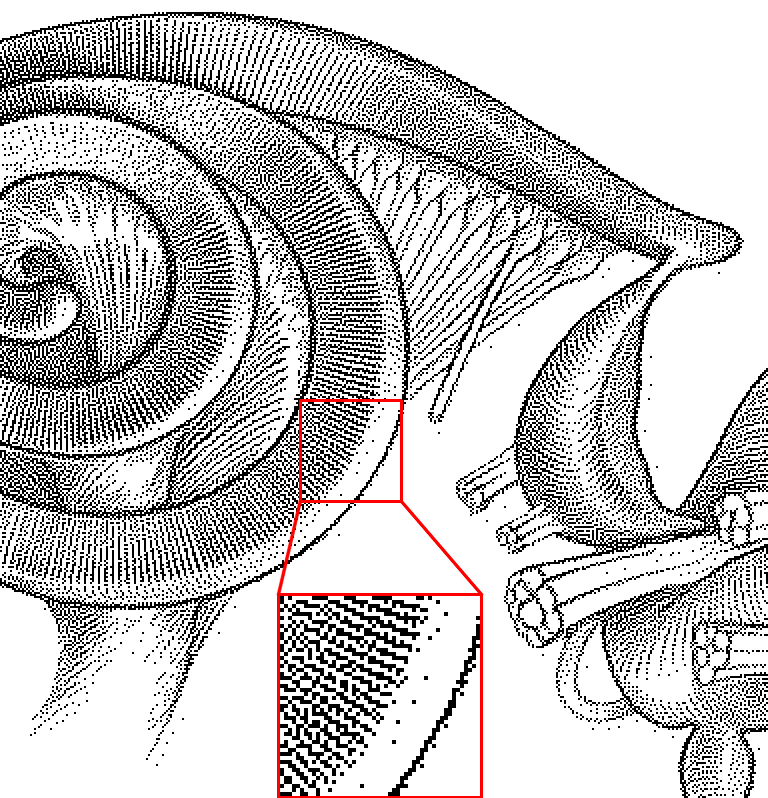}}
  \vfill
  \vspace{-2mm}
  \subfloat[\scriptsize{$\text{TDED}_\text{BS}$ \cite{fung2016}}]{\includegraphics[width=0.24\linewidth]{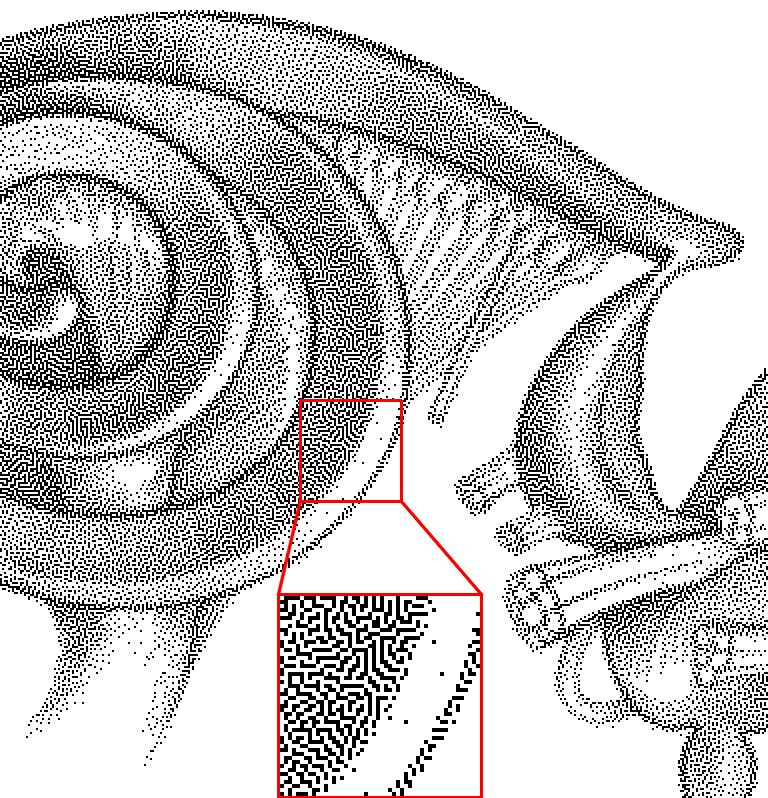}}
  \hspace{1mm}
  \subfloat[\scriptsize{SGED \cite{hu2016}}]{\includegraphics[width=0.24\linewidth]{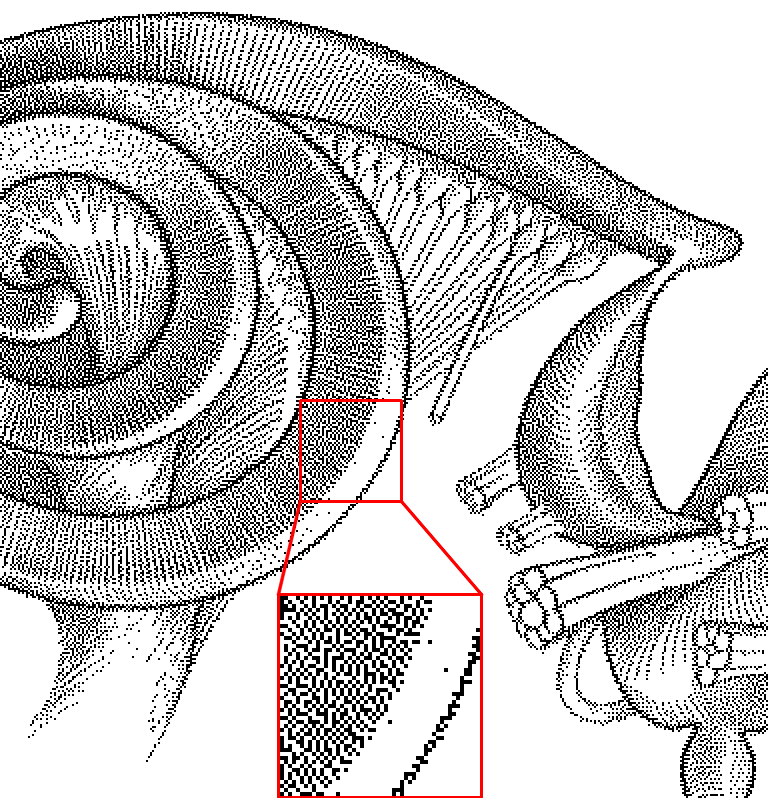}}
  \hspace{1mm}
  \subfloat[\scriptsize{DBS \cite{analoui1992}}]{\includegraphics[width=0.24\linewidth]{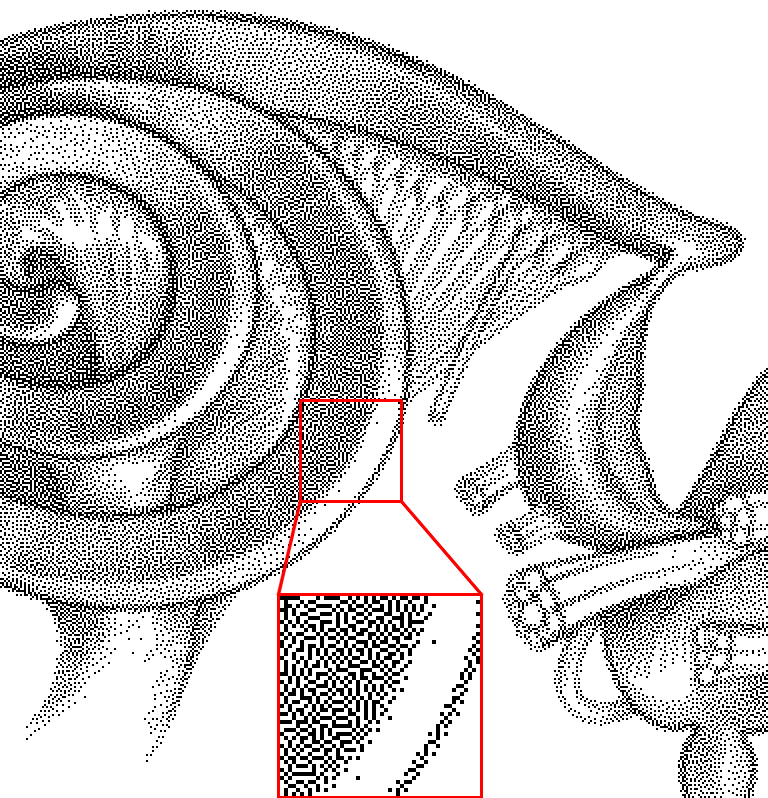}}
  \hspace{1mm}
  \subfloat[\scriptsize{SAH \cite{pang2008}}]{\includegraphics[width=0.24\linewidth]{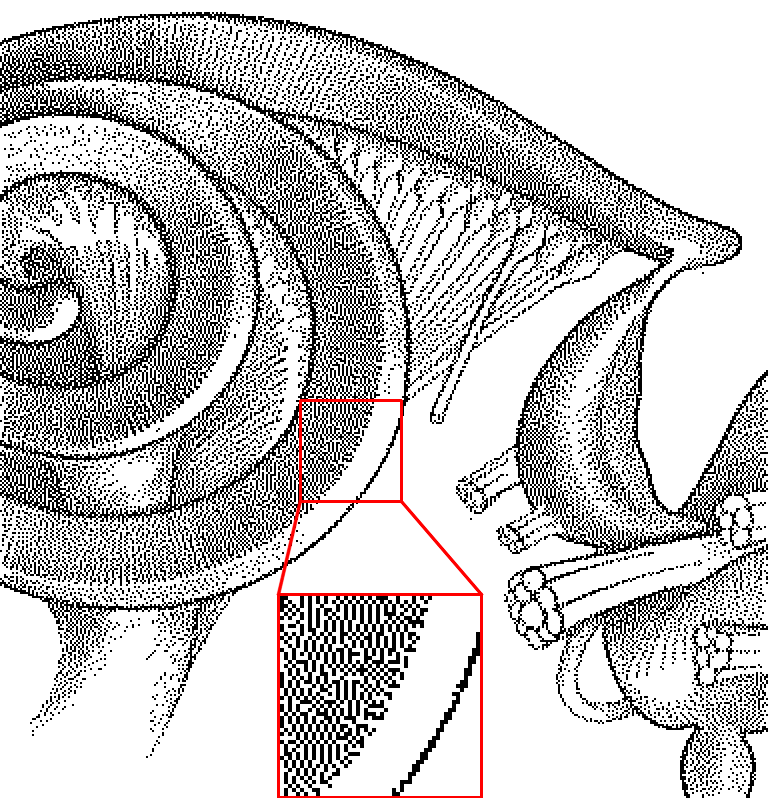}}
  \vfill
  \vspace{-2mm}
  \subfloat[\scriptsize{RVH \cite{xia2021} w/ recons.}]{\includegraphics[width=0.24\linewidth]{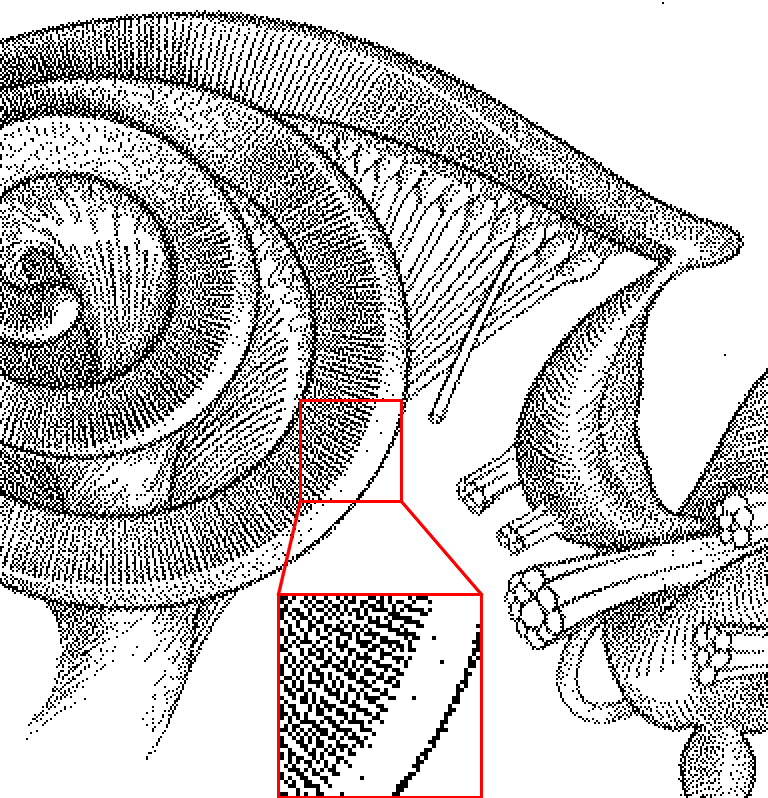}}
  \hspace{1mm}
  \subfloat[\scriptsize{RVH \cite{xia2021} w/o recons.}]{\includegraphics[width=0.24\linewidth]{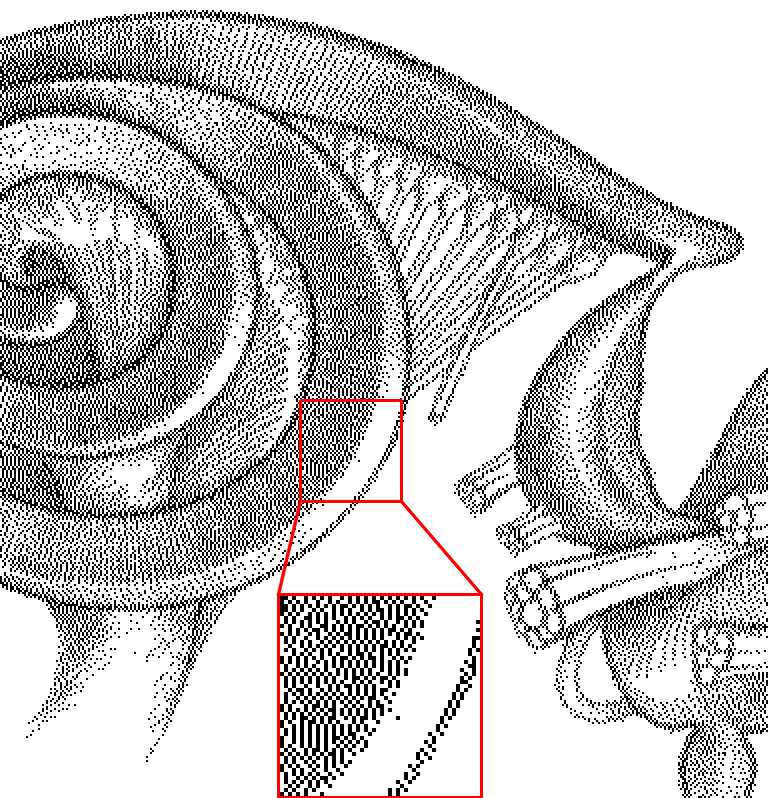}}
  \hspace{1mm}
  \subfloat[\scriptsize{cGAN \cite{choi2022}}]{\includegraphics[width=0.24\linewidth]{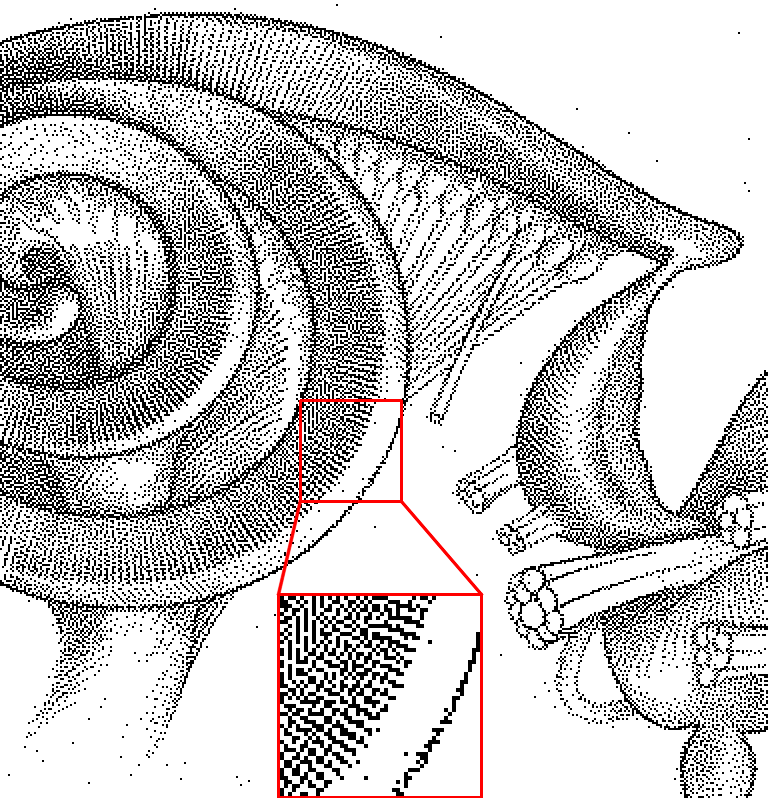}}
  \hspace{1mm}
  \subfloat[\scriptsize{Ours}]{\includegraphics[width=0.24\linewidth]{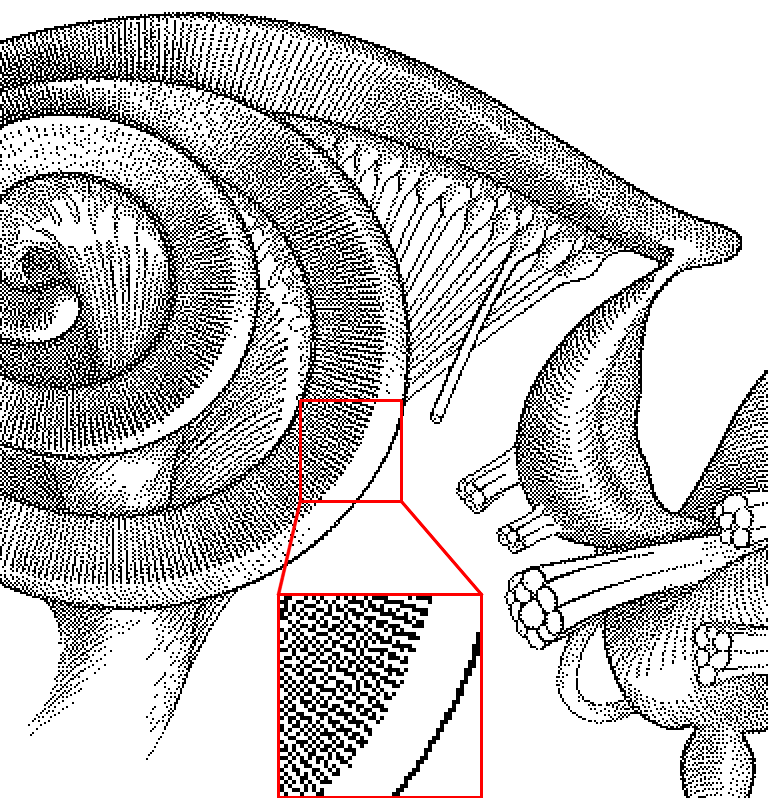}}
  \caption{Halftone samples of image ``Snail Shaped Organ''.}
  \label{fig:visual_compare_1}
\end{figure*}

\begin{figure*}[t]
  \centering
  \captionsetup[subfloat]{labelsep=none,format=plain,labelformat=empty}
  \subfloat[\scriptsize{CT}]{\includegraphics[width=0.24\linewidth]{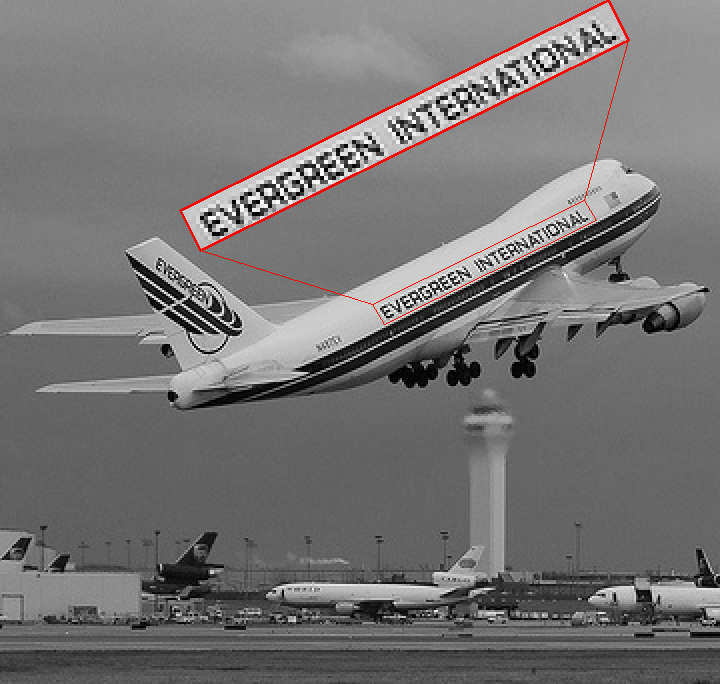}}
  \hspace{1mm}
  \subfloat[\scriptsize{VAC \cite{ulichney1993}}]{\includegraphics[width=0.24\linewidth]{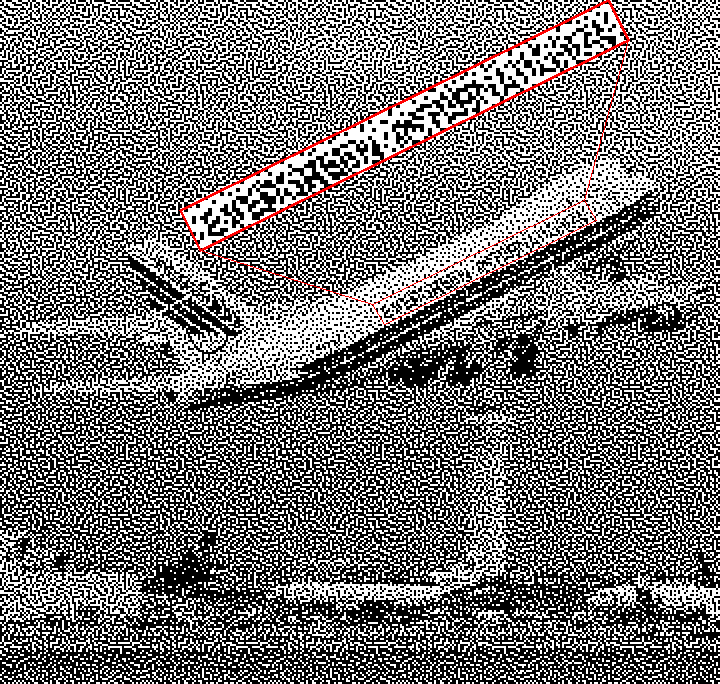}}
  \hspace{1mm}
  \subfloat[\scriptsize{OVED \cite{ostromoukhov2001}}]{\includegraphics[width=0.24\linewidth]{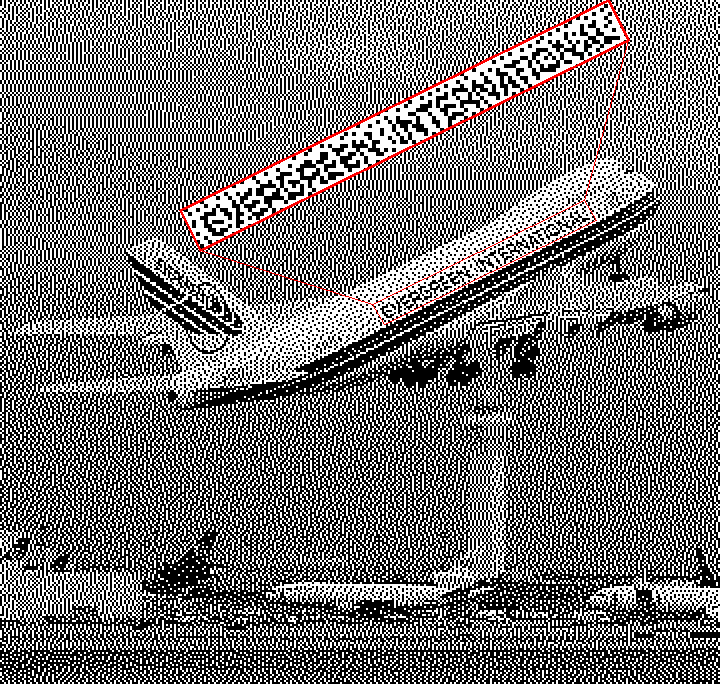}}
  \hspace{1mm}
  \subfloat[\scriptsize{SAED \cite{chang2009}}]{\includegraphics[width=0.24\linewidth]{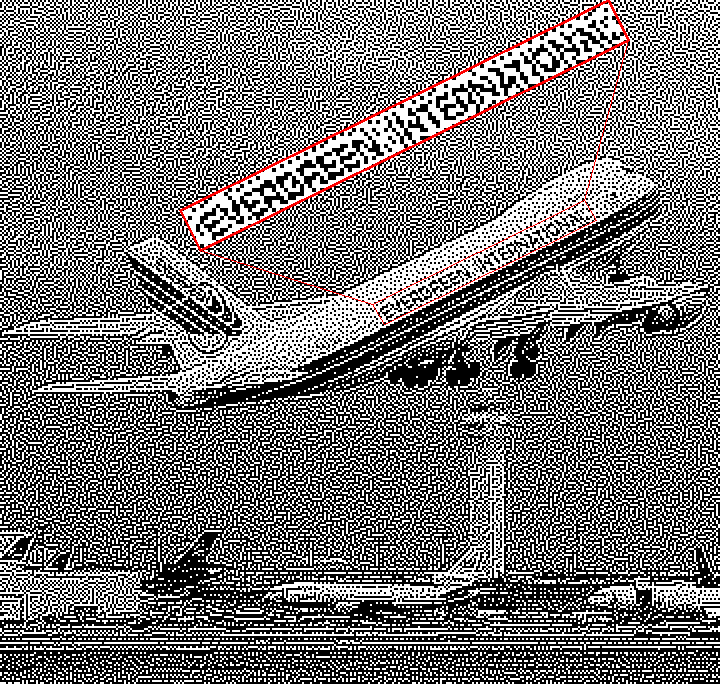}}
  \vfill
  \vspace{-2mm}
  \subfloat[\scriptsize{$\text{TDED}_\text{BS}$ \cite{fung2016}}]{\includegraphics[width=0.24\linewidth]{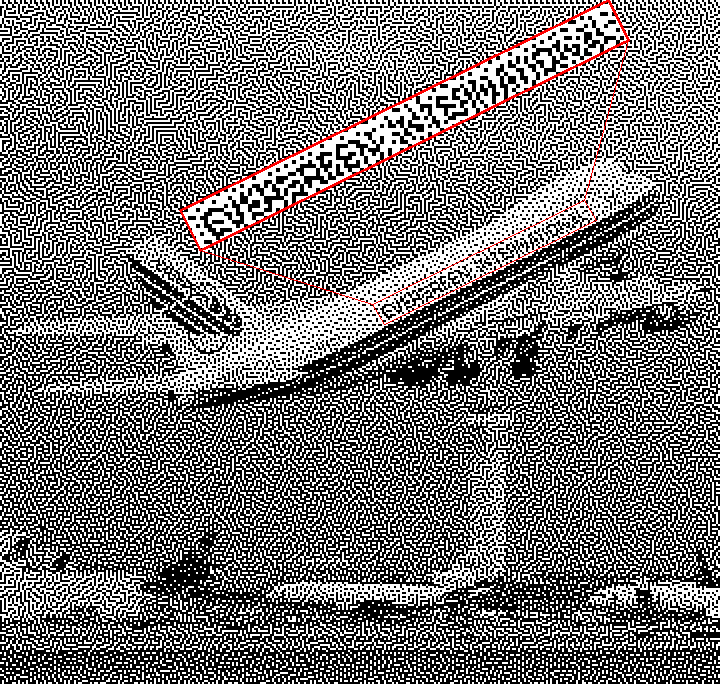}}
  \hspace{1mm}
  \subfloat[\scriptsize{SGED \cite{hu2016}}]{\includegraphics[width=0.24\linewidth]{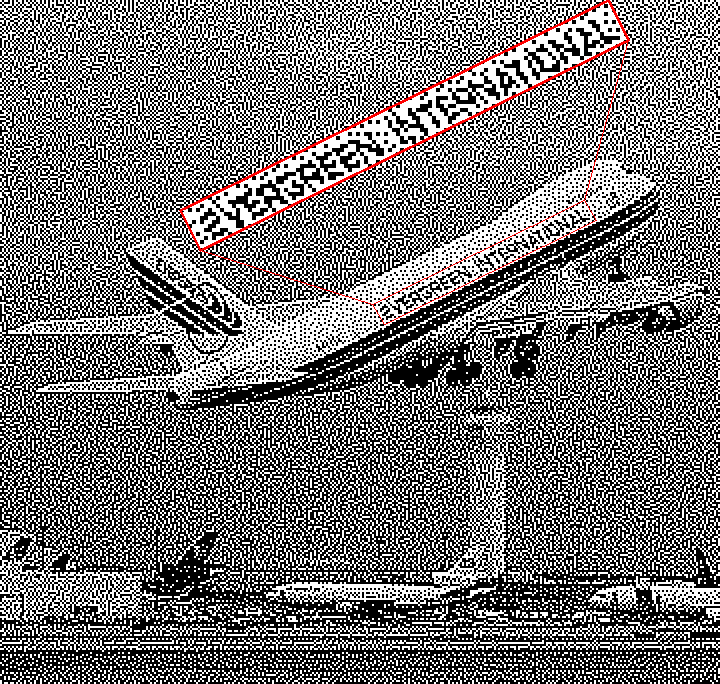}}
  \hspace{1mm}
  \subfloat[\scriptsize{DBS \cite{analoui1992}}]{\includegraphics[width=0.24\linewidth]{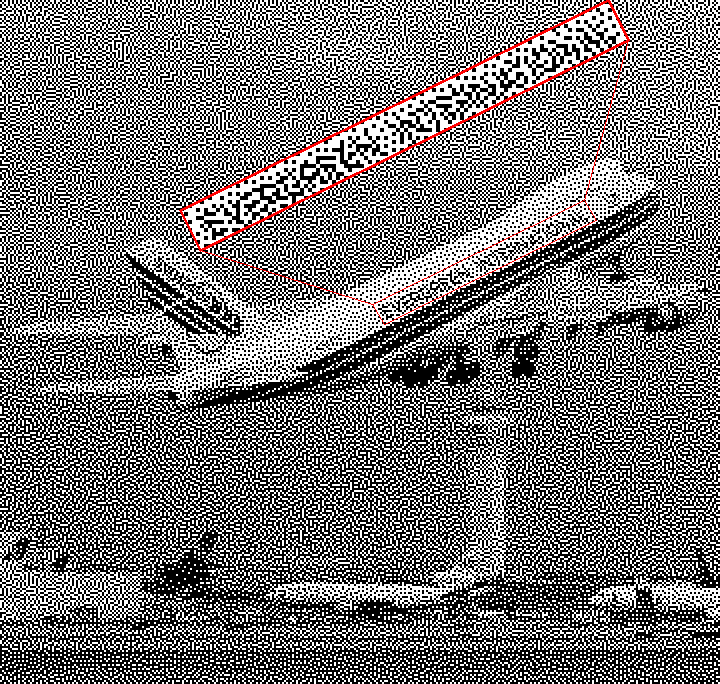}}
  \hspace{1mm}
  \subfloat[\scriptsize{SAH \cite{pang2008}}]{\includegraphics[width=0.24\linewidth]{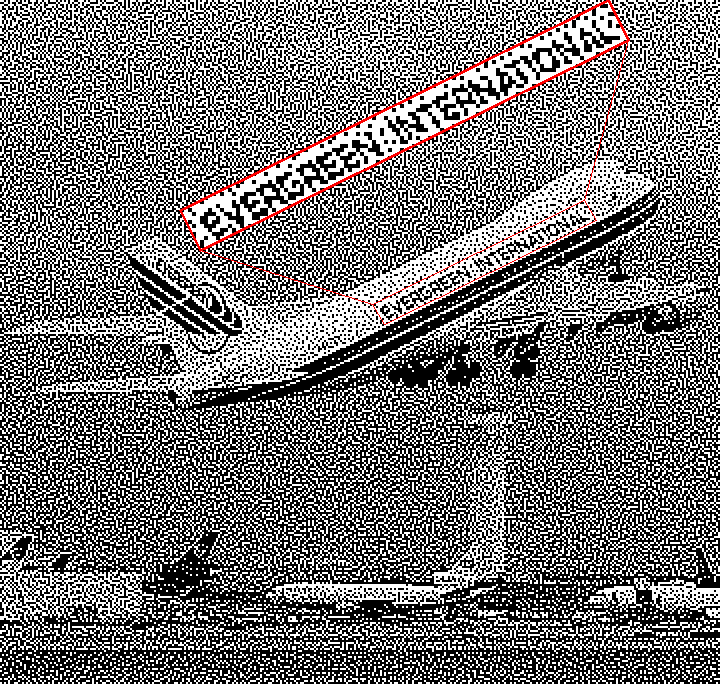}}
  \vfill
  \vspace{-2mm}
  \subfloat[\scriptsize{RVH \cite{xia2021} w/ recons.}]{\includegraphics[width=0.24\linewidth]{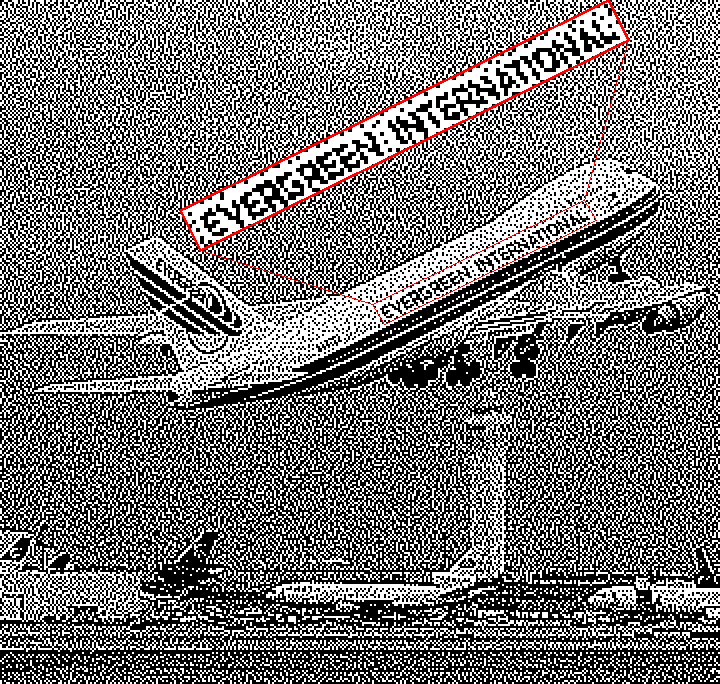}}
  \hspace{1mm}
  \subfloat[\scriptsize{RVH \cite{xia2021} w/o recons.}]{\includegraphics[width=0.24\linewidth]{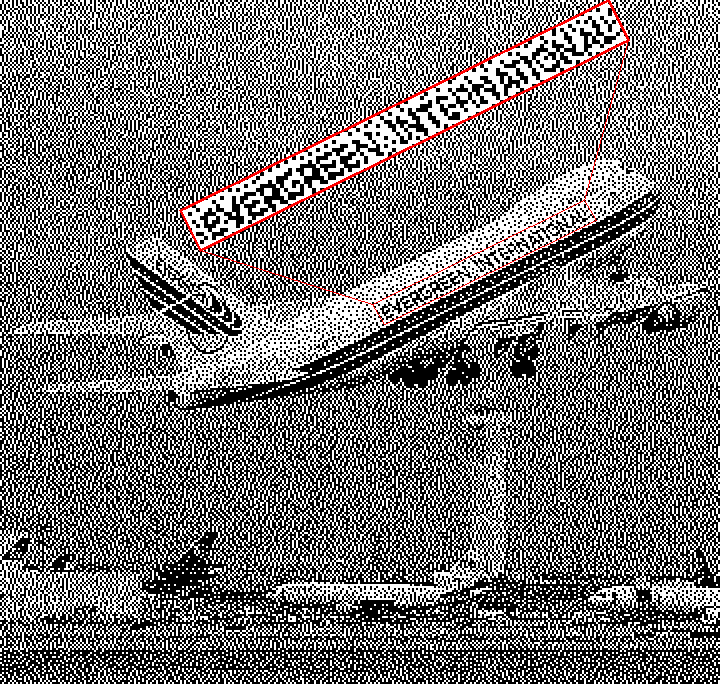}}
  \hspace{1mm}
  \subfloat[\scriptsize{cGAN \cite{choi2022}}]{\includegraphics[width=0.24\linewidth]{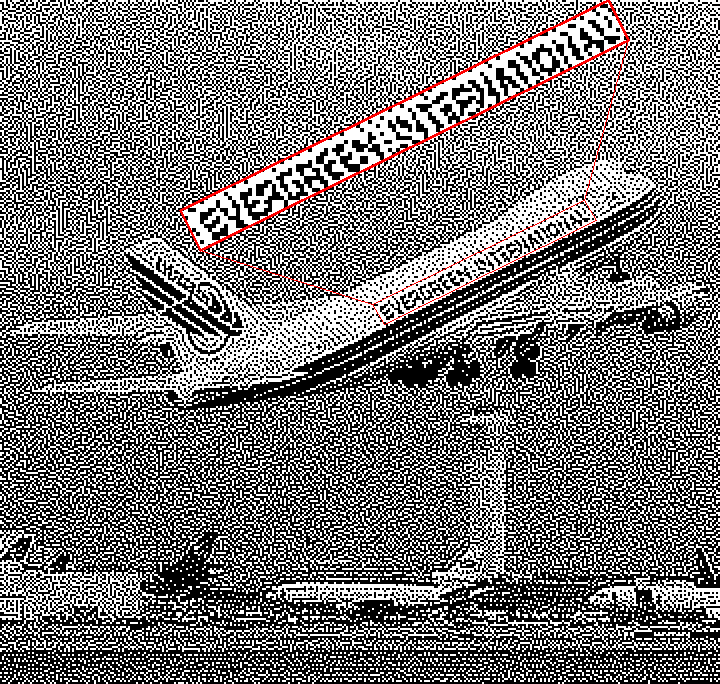}}
  \hspace{1mm}
  \subfloat[\scriptsize{Ours}]{\includegraphics[width=0.24\linewidth]{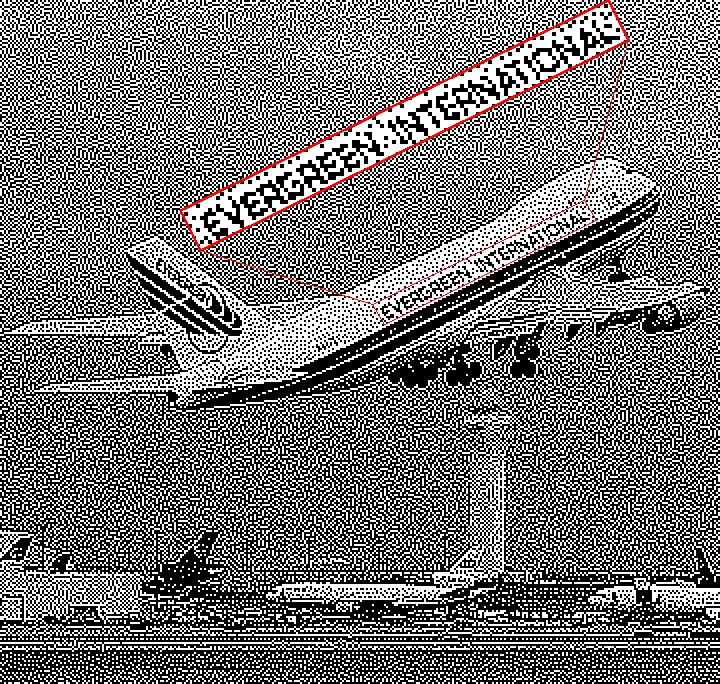}}
  \caption{Halftone samples of image ``Plane'' (from VOC2012 dataset \cite{voc2012}).}
  \label{fig:visual_compare_2}
\end{figure*}

\textbf{Dataset.} We evaluate all halftoning methods on the VOC2012 dataset \cite{voc2012} following \cite{xia2021}.
The test set contains 1,684 images, which are randomly selected from 17,125 images.
For learning-based methods (RVH, cGAN and ours), the remaining 13,758 and 1,683 images are reserved as the training set and the validation set, respectively.
All images are converted to grayscale in advance.
Note that our formulation can be considered as self-supervised, so it should be easy to collect more label-free continuous-tone images.

\textbf{HW/SW Environment.} To take full advantage of the parallelism in some halftoning methods (VAC, SAED, RVH, cGAN, and ours), we implement them on an NVIDIA GeForce RTX 2080Ti GPU with PyTorch 1.11.0, CUDA 11.3 and cuDNN 8.2.0.
For those serial methods (OVED, $\text{TDED}_\text{BS}$, SGED, DBS and SAH), we implement them in C++ with GCC 6.3.0 and OpenCV 4.6.0 on an Intel Xeon Gold 5215 CPU (2.5GHz).

\textbf{Details of Our Method.}
We choose ResNet \cite{he2016} as the backbone policy network.
The CNN has 16 residual blocks and 33 convolutional layers in total.
Each convolution kernel has 32 channels, and the strides of all convolution layers are 1.
A pixel-wise sigmoid layer has been appended to the CNN to output the probabilities of white dots $\bpi(\bh=\bone)$.
All convolution kernels' weights are initialized by sampling from the normal distribution $\mathcal{N}(0, 0.01^2)$, and the biases are initialized to zero.
This is to set the initial action probabilities around 0.5, so agents can do sufficient exploration rather than be stuck in one random action preference.
The batch size is set to 64, training samples are randomly cropped to 64x64, hyper-parameters $w_s = 0.06$, $w_a = 0.002$, and the learning rate $\alpha$ is adjusted from 3e-4 to 1e-5 with the cosine annealing schedule.
The model is trained for 200,000 iterations ($\sim$12 hours on a 2080Ti GPU) with the Adam optimizer \cite{kingma2015}.
In the testing phase, we process and evaluate images with their original size.
We do not pad the images when calculating metrics.

\subsection{Halftone Quality}

\begin{figure}[t]
  \begin{minipage}[b]{\linewidth}
    \centering
    \begin{overpic}[width=\linewidth]{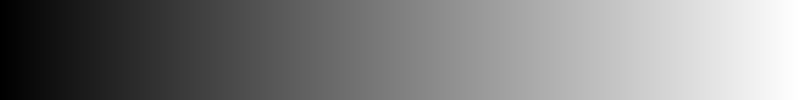}
      \put(92, 1){\footnotesize\colorbox{white}{CT}}
    \end{overpic}
  \end{minipage}
  \vfill \vspace{1mm}
  \begin{minipage}[b]{\linewidth}
    \centering
    \begin{overpic}[width=\linewidth]{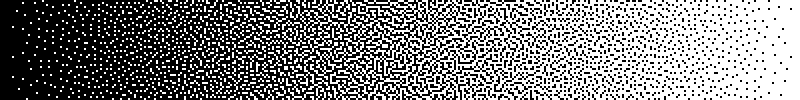}
      \put(90, 1){\footnotesize\colorbox{white}{VAC}}
    \end{overpic}
  \end{minipage}
  \vfill \vspace{1mm}
  \begin{minipage}[b]{\linewidth}
    \centering
    \begin{overpic}[width=\linewidth]{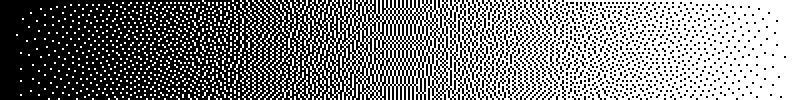}
      \put(88, 1){\footnotesize\colorbox{white}{OVED}}
    \end{overpic}
  \end{minipage}
  \vfill \vspace{1mm}
  \begin{minipage}[b]{\linewidth}
    \centering
    \begin{overpic}[width=\linewidth]{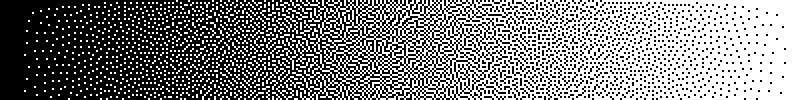}
      \put(88, 1){\footnotesize\colorbox{white}{SAED}}
    \end{overpic}
  \end{minipage}
  \vfill \vspace{1mm}
  \begin{minipage}[b]{\linewidth}
    \centering
    \begin{overpic}[width=\linewidth]{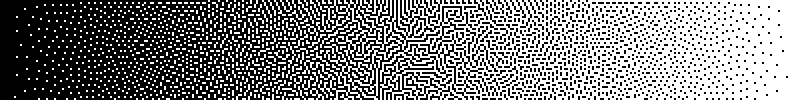}
      \put(85, 1){\footnotesize\colorbox{white}{$\text{TDED}_\text{BS}$}}
    \end{overpic}
  \end{minipage}
  \vfill \vspace{1mm}
  \begin{minipage}[b]{\linewidth}
    \centering
    \begin{overpic}[width=\linewidth]{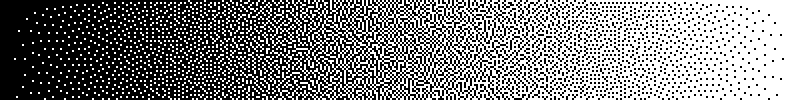}
      \put(88, 1){\footnotesize\colorbox{white}{SGED}}
    \end{overpic}
  \end{minipage}
  \vfill \vspace{1mm}
  \begin{minipage}[b]{\linewidth}
    \centering
    \begin{overpic}[width=\linewidth]{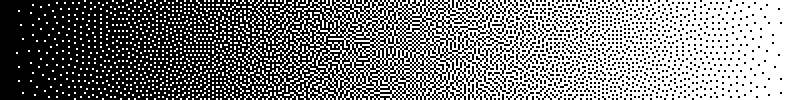}
      \put(90, 1){\footnotesize\colorbox{white}{DBS}}
    \end{overpic}
  \end{minipage}
  \vfill \vspace{1mm}
  \begin{minipage}[b]{\linewidth}
    \centering
    \begin{overpic}[width=\linewidth]{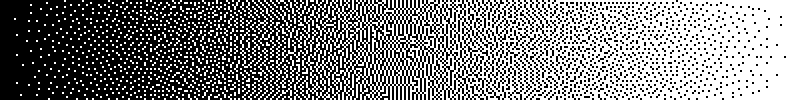}
      \put(90, 1){\footnotesize\colorbox{white}{SAH}}
    \end{overpic}
  \end{minipage}
  \vfill \vspace{1mm}
  \begin{minipage}[b]{\linewidth}
    \centering
    \begin{overpic}[width=\linewidth]{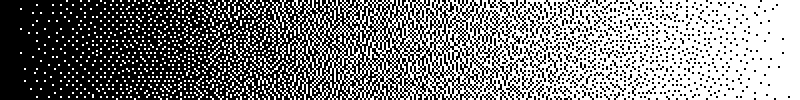}
      \put(74, 1){\footnotesize\colorbox{white}{RVH w/o recons.}}
    \end{overpic}
  \end{minipage}
  \vfill \vspace{1mm}
  \begin{minipage}[b]{\linewidth}
    \centering
    \begin{overpic}[width=\linewidth]{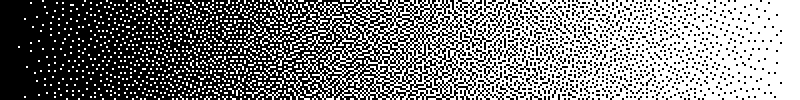}
      \put(88, 1){\footnotesize\colorbox{white}{cGAN}}
    \end{overpic}
  \end{minipage}
  \vfill \vspace{1mm}
  \begin{minipage}[b]{\linewidth}
    \centering
    \begin{overpic}[width=\linewidth]{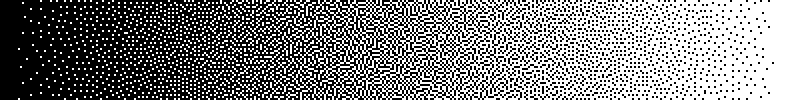}
      \put(90, 1){\footnotesize\colorbox{white}{Ours}}
    \end{overpic}
  \end{minipage}
  \vfill
  \caption{Halftone samples of grayscale intensity ramp.}
  \label{fig:ramps}
\end{figure}

To compare various methods' visual quality, we test them on different kinds of images and show three samples here.

Fig.~\ref{fig:visual_compare_1} shows the benefit of effectively optimizing halftone metric by our method.
Our halftone results keep identifiable structures from the source image.
On the contrary, VAC, OVED, $\text{TDED}_\text{BS}$ and DBS totally destroy the vein pattern because they do not take the structural similarity into consideration.
With regard to RVH, we find the halftone with reversibility has better structural details than the one without considering reconstruction.
We think this is because the structures have to be maintained in the halftone for the need of reconstruction.
The results obtained from cGAN lack structural details, as the dataset used and the loss function employed do not take into account this information.
SAED successfully generates halftones with rich structural features.
However, there are isolated minority dots near the edges in SAED's result.
This is due to the fact that the accumulated errors are improperly released here by lack of homogeneous receivers.
The SAH method, which needs per-instance tuning \cite{xia2021}, shows unsatisfactory results in our experiment with the default parameters in \cite{pang2008}.

Fig.~\ref{fig:visual_compare_2} shows a realistic image and its halftone results.
Among all the existing methods, SAED achieves the relatively good quality.
However, the text on the plane is not very clear, which can be blamed on the assumption of only one dominant frequency at each position \cite{chang2009}.
In contrast, our method is more adaptive and can render legible structures by learning to optimize the proposed CSSIM metric.
Note that all methods except $\text{TDED}_\text{BS}$ have shown a certain degree of checkerboard artifacts in the mid-tone area.
In $\text{TDED}_\text{BS}$, a better noise model has been explicitly considered.
For more details of the updated noise model, we refer to \cite{lau2006,fung2013,fung2016}.

For reference, Fig.~\ref{fig:ramps} shows the results of all methods on the grayscale ramp image.
The quality of our halftone result is comparable to the DBS method.

\begin{table*}[t]
  \caption{Quantitative comparison of halftoning methods. Metrics are measured on the VOC2012 test set. Runtimes are measured on the 512x512 ``Lenna'' test image. ``*'' indicates this method is GPU-accelerated.\label{tab:quant_results}}
  \centering
  \begin{tabular}{cc|cccc|ccc}
    \toprule
    \textbf{Method}                         &               & \textbf{PSNR} (Näsänen)   & \textbf{PSNR} (Gaussian)  & \textbf{SSIM} \cite{wang2004} & \textbf{CSSIM} (Eq.~\eqref{eqn:cssim}) & \textbf{Param. \#}     & \textbf{Runtime}          \\
    \midrule \midrule
    VAC \cite{ulichney1993}                 & (64x64)       & 29.191$\pm$0.836          & 34.472$\pm$1.758          & 0.0808$\pm$0.0584             & 0.9075$\pm$0.0370                      & 4K                     & 0.03ms*                   \\
    VAC \cite{ulichney1993}                 & (512x512)     & 29.283$\pm$0.839          & 34.490$\pm$1.764          & 0.0808$\pm$0.0584             & 0.9075$\pm$0.0370                      & 262K                   & 0.03ms*                   \\
    \midrule
    OVED \cite{ostromoukhov2001}            &               & 33.052$\pm$0.481          & 42.795$\pm$1.259          & 0.0998$\pm$0.0733             & 0.9117$\pm$0.0339                      & 384                    & 2.36ms                    \\
    SAED \cite{chang2009}                   &               & 30.942$\pm$0.831          & 37.786$\pm$1.683          & 0.1220$\pm$0.0774             & 0.9175$\pm$0.0302                      & 341K                   & 435.25ms*                 \\
    $\text{TDED}_\text{BS}$ \cite{fung2016} &               & 31.798$\pm$0.519          & 42.419$\pm$0.521          & 0.0887$\pm$0.0679             & 0.9085$\pm$0.0363                      & 818                    & 3.07ms                    \\
    SGED \cite{hu2016}                      &               & 31.441$\pm$0.675          & 38.360$\pm$1.163          & 0.1391$\pm$0.0849             & 0.9195$\pm$0.0289                      & -                      & 9.87ms                    \\
    \midrule
    DBS \cite{analoui1992}                  &               & \textbf{33.987}$\pm$0.414 & \textbf{44.365}$\pm$0.534 & 0.0816$\pm$0.0675             & 0.9055$\pm$0.0389                      & -                      & 932.49ms                  \\
    SAH \cite{pang2008}                     &               & 31.214$\pm$1.095          & 38.018$\pm$2.160          & 0.1380$\pm$0.0934             & 0.9210$\pm$0.0275                      & -                      & 162s                      \\
    \midrule
    RVH \cite{xia2021}                      & (w/ recons.)  & 29.447$\pm$0.737          & 34.956$\pm$1.074          & 0.1500$\pm$0.0912             & 0.9220$\pm$0.0272                      & \multirow{2}{*}{37.8M} & \multirow{2}{*}{49.61ms*} \\
    RVH \cite{xia2021}                      & (w/o recons.) & 30.573$\pm$0.562          & 34.658$\pm$0.700          & 0.1360$\pm$0.0907             & 0.9172$\pm$0.0298                      &                        &                           \\
    cGAN \cite{choi2022}                    &               & 30.540$\pm$0.553          & 36.167$\pm$0.500          & 0.1133$\pm$0.0770             & 0.9159$\pm$0.0316                      & 3.1M                   & 166.67ms*                 \\
    \textbf{Ours}                           &               & 31.642$\pm$0.670          & 37.547$\pm$0.813          & \textbf{0.1610}$\pm$0.0959    & \textbf{0.9236}$\pm$0.0264             & 299K                   & 28.38ms*                  \\
    \bottomrule
  \end{tabular}
\end{table*}

Quantitative quality results of all methods on the test dataset are listed in Table~\ref{tab:quant_results} (mean and standard deviation).
The tone consistency is measured by PSNR between the HVS-filtered halftone and the HVS-filtered input contone.
In addition to the Näsänen HVS model used in our work, we also report the PSNR scores calculated with the Gaussian filter (sigma=2) for reference.
The structural similarity is measured by the SSIM\footnote{Note that it is not appropriate to assess halftones directly using SSIM \cite{itoua2010}. We list SSIM scores here for reference.} and the proposed CSSIM (Eq.~\eqref{eqn:cssim}).
Thanks to the effective DRL framework, our model achieves the best SSIM and CSSIM scores among all candidates.
Surprisingly, it performs even better than the search-based SAH.
With limited searching budgets, it seems hard for SAH to effectively optimize the target with a generic heuristic searching strategy.
With regard to PSNR, DBS achieves the best score by effectively searching \cite{liao2015} for an optimal halftone solution of one contone image on-the-fly.
Although halftones with an extremely high PSNR are not necessarily more visually pleasing, it is possible to narrow the gap between DBS and our method by decreasing $w_s$ or increasing the network capacity.

\subsection{Processing Efficiency}
The parameter number (predefined dither array, look up tables or trained weights) and the runtime (evaluated on the 512x512 ``Lenna'' testcase) of all methods are listed in Table~\ref{tab:quant_results}.
While the ordered dithering method (VAC) is the fastest, its halftone results are of inferior quality.
However, this strategy is still useful in situations where the quality requirements are not strict.
Search-based methods (DBS and SAH) are quite time-consuming which impedes its practical application.
Compared to them, SAED achieves a better balance between speed and quality.
The SGED method is competitive, although it shows limited quality in some complex scenarios (such Fig.~\ref{fig:visual_compare_1}).
RVH's, cGAN's and our results prove that it is possible to learn a halftoning network and leave the time-consuming optimization in the offline phase.
Furthermore, our method can enable a light-weight CNN to perform halftoning with better image quality.

\subsection{Discussions}
\textbf{Comparison of Gradient Estimators.}
To make the quantized image differentiable with respect to the $\theta$, Yoo \etal \cite{yoo2020} introduced the soft projection operation that relaxes the quantization with a small temperature.
Later, Xia \etal \cite{xia2021} found that a simple relaxation alone was not sufficient for halftones, which has only two discrete levels.
They proposed the binarization loss that greedily pushes the predictions to their nearest discrete levels.
Besides, rather than stochastic policies, both \cite{yoo2020} and \cite{xia2021} utilized deterministic models $\bh = \text{Thresholding}(\text{CNN}_\theta(\bc))$.
Since the gradient of the $\text{Thresholding}(\cdot)$ operation vanishes almost everywhere, which impedes the training, the STE \cite{bengio2013} that assumes $\nabla_\theta \bh \approx \nabla_\theta \text{CNN}_\theta(\bc,\bz)$ was also tested by \cite{xia2021}.
Nevertheless, this approximated gradient is generally not the gradient of any function \cite{yin2019}.
On the contrary, the unbiased $\hat{g}_\text{COMA}$ (Eq.~\eqref{eqn:coma}) and $\hat{g}_\text{LE}$ (Eq.~\eqref{eqn:le}) are directly derived from the target (Eq.~\eqref{eqn:gradient}), which inherently do optimization and discretization at the same time.

\begin{figure}[t]
  \centering
  \includegraphics[width=0.9\linewidth]{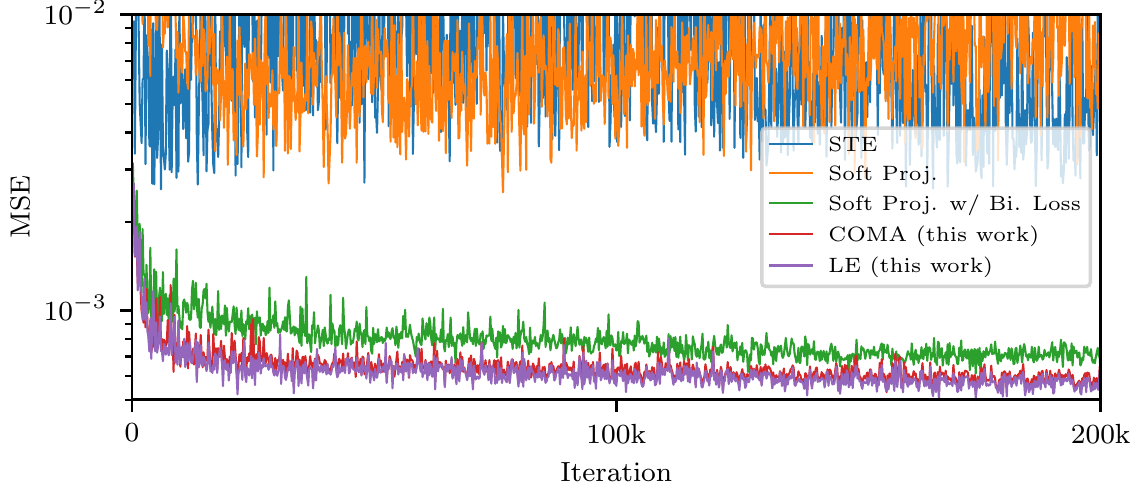}
  \vspace{-3mm}
  \caption{MSE training loss curves of gradient estimators.}
  \label{fig:training_curves}
\end{figure}

\begin{figure}[t]
  \centering
  \includegraphics[width=0.8\linewidth]{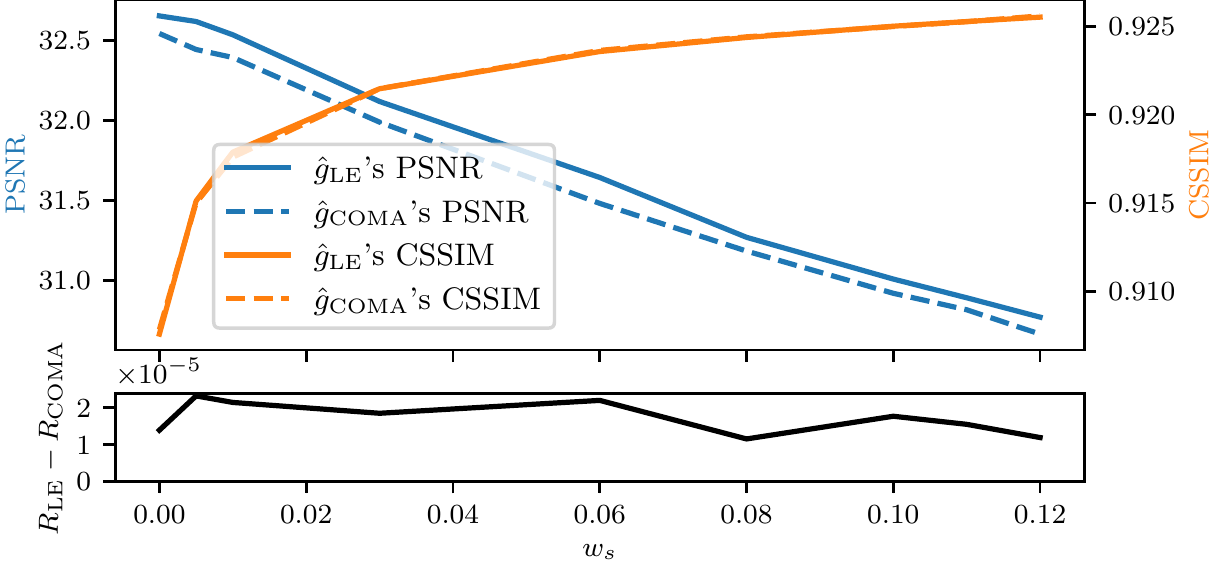}
  \caption{Optimizing results (PSNR and CSSIM) by $\hat{g}_{\text{LE}}$ and $\hat{g}_{\text{COMA}}$ with different $w_s$, as well as their performance comparison $R_\text{LE} - R_\text{COMA}$.}
  \label{fig:w_s_sweep}
\end{figure}

To test the effectiveness of these gradient estimators, we conduct training experiments using the same CNN model and optimization objective as in our method.
The MSE loss curves are plotted in Fig.~\ref{fig:training_curves}.
We find that, firstly, simply adopting the STE or relaxation neither decreases MSE nor renders blue-noise halftones.
Second, while the binarization loss proposed in \cite{xia2021} does help cluster the outputs, this greedy rule could harm the optimization (PSNR (Näsänen/Gaussian), SSIM and CSSIM scores: 30.749/35.974/0.1529/0.9229).
By contrast, our estimators $\hat{g}_{\text{LE}}$ and $\hat{g}_{\text{COMA}}$ result in stable and quick optimization processes.
Last but not least, we find that $\hat{g}_{\text{LE}}$ performs better than $\hat{g}_{\text{COMA}}$ on PSNR (see Fig.~\ref{fig:w_s_sweep}) while their computational costs are the same.

\begin{figure}[t]
  \centering
  \begin{minipage}{\linewidth}
    \centering
    \begin{minipage}{0.32\linewidth}
      \centering
      \includegraphics[width=\linewidth]{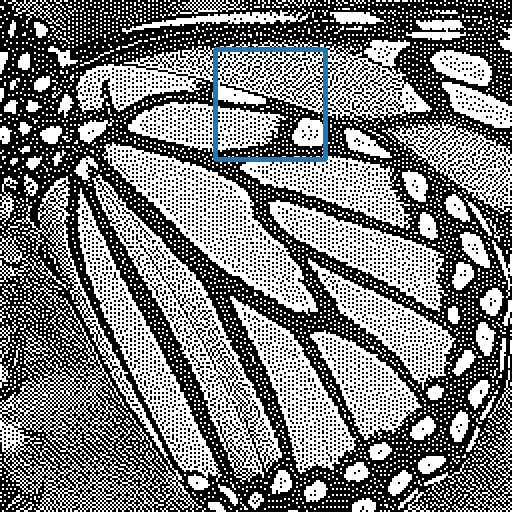}
    \end{minipage}
    \begin{minipage}{0.32\linewidth}
      \centering
      \includegraphics[width=\linewidth]{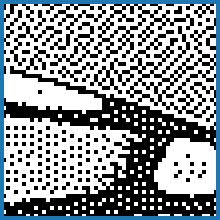}
    \end{minipage}
    \begin{minipage}{0.32\linewidth}
      \centering
      \includegraphics[width=\linewidth]{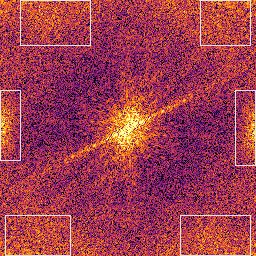}
    \end{minipage}
    \vfill
    \vspace{0.5mm}
    \small{(a)}
    \vfill
    \vspace{0.5mm}
  \end{minipage}
  \vfill
  \begin{minipage}{\linewidth}
    \centering
    \begin{minipage}{0.32\linewidth}
      \centering
      \includegraphics[width=\linewidth]{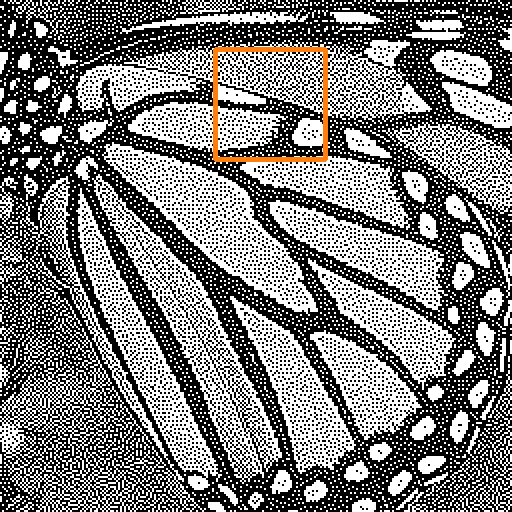}
    \end{minipage}
    \begin{minipage}{0.32\linewidth}
      \centering
      \includegraphics[width=\linewidth]{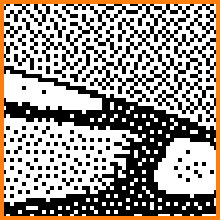}
    \end{minipage}
    \begin{minipage}{0.32\linewidth}
      \centering
      \includegraphics[width=\linewidth]{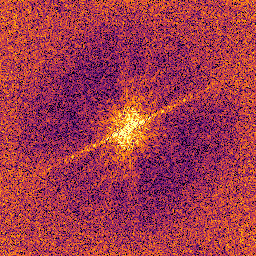}
    \end{minipage}
    \vfill
    \vspace{0.3mm}
    \small{(b)}
  \end{minipage}
  \caption{Halftone samples of image ``Butterfly'' with Fourier amplitude spectrum: (a) w/o $\mathcal{L}_{AS}$ ($w_a=0$). (b) w/ $\mathcal{L}_{AS}$ ($w_a=0.002$).}
  \label{fig:as_ablation}
\end{figure}

\begin{figure}[t]
  \centering
  \begin{minipage}{0.48\linewidth}
    \centering
    \renewcommand{\arraystretch}{1.2}
    \resizebox{\linewidth}{!}{
      \begin{tabular}{|c|c|c|}
        \hline
        Noise Type                        & PSNR   & CSSIM  \\
        \hline
        $\mathcal{N}(0,1)$                & 31.642 & 0.9236 \\
        $\mathcal{U}(0,1)$                & 31.663 & 0.9236 \\
        $\Gamma(1,10)$                    & 31.636 & 0.9236 \\
        $\mathrm{B}(0.5)$                 & 31.573 & 0.9237 \\
        Position Encoding \cite{kim2018}  & 31.685 & 0.9237 \\
        Dither Array \cite{ulichney1993}  & 31.644 & 0.9236 \\
        Screened Halftone \cite{choi2022} & 31.616 & 0.9237 \\
        \hline
      \end{tabular}
    }
  \end{minipage}
  \begin{minipage}{0.42\linewidth}
    \centering
    \includegraphics[width=\linewidth]{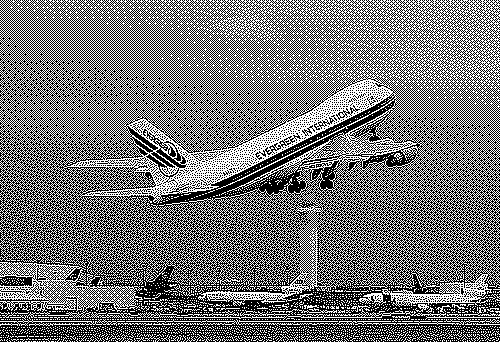}
  \end{minipage}
  \caption{Left: optimizing results with different input noise type. Right: the ``Plane'' halftone sample generated with position encoding noise \cite{kim2018}.}
  \label{fig:noise_type}
\end{figure}

\begin{figure*}[t]
  \centering
  \begin{minipage}{0.2\linewidth}
    \centering
    \includegraphics[width=\linewidth]{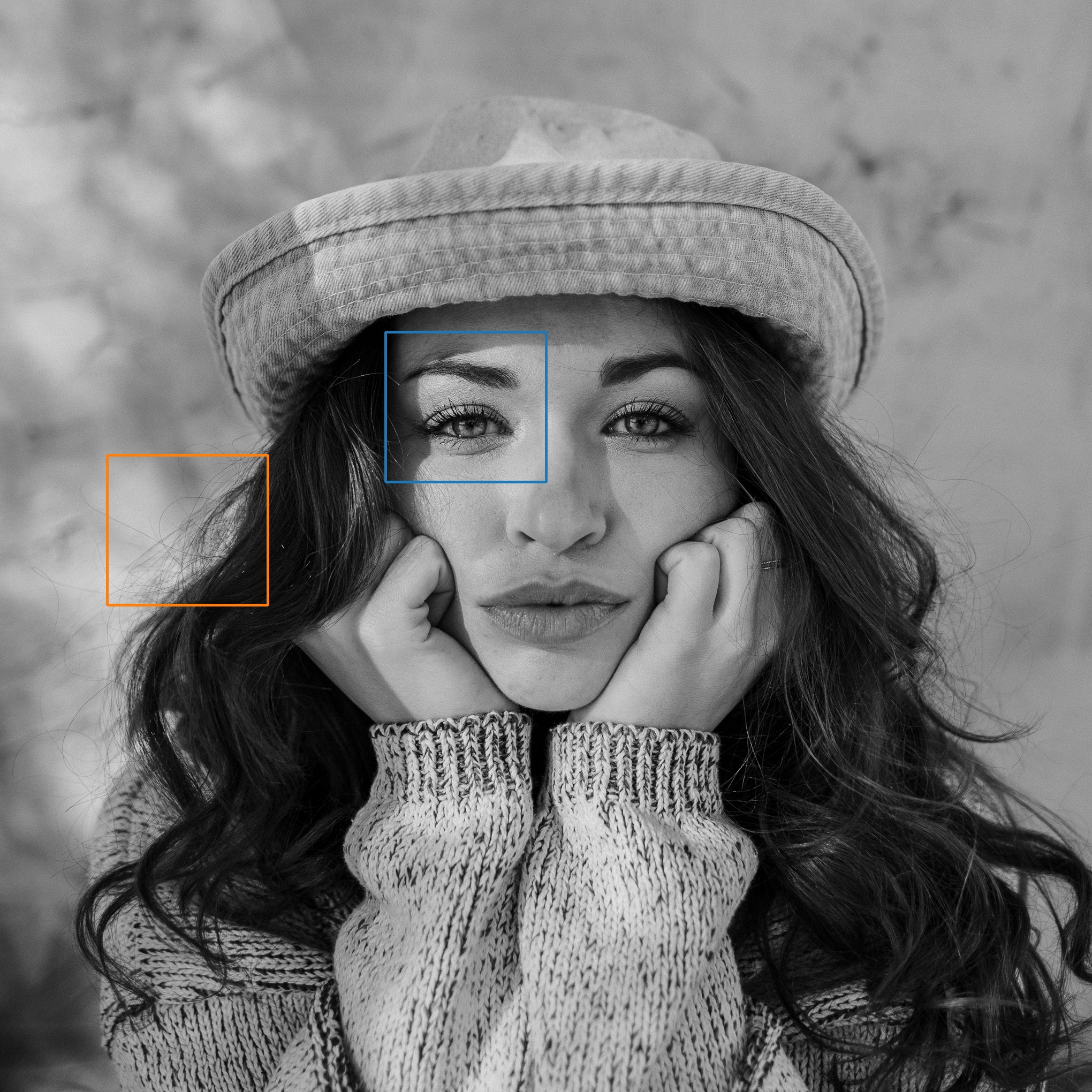}
    \vfill
    \small{CT}
  \end{minipage}
  \hfill
  \begin{minipage}{0.78\linewidth}
    \centering
    \begin{minipage}{0.24\linewidth}
      \centering
      \includegraphics[width=\linewidth]{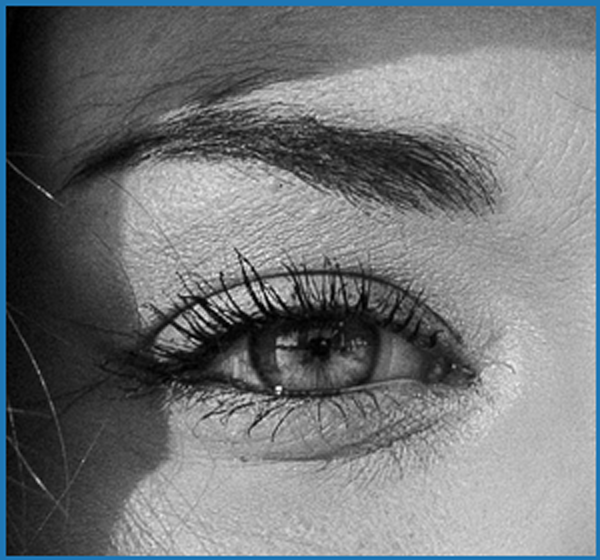}
      \vfill
      \vspace{0.5mm}
      \includegraphics[width=\linewidth]{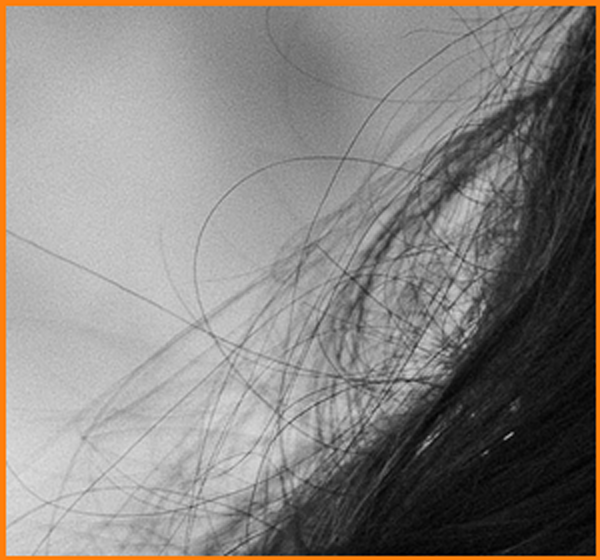}
      \vfill
      \small{CT}
    \end{minipage}
    \hfill
    \begin{minipage}{0.24\linewidth}
      \centering
      \includegraphics[width=\linewidth]{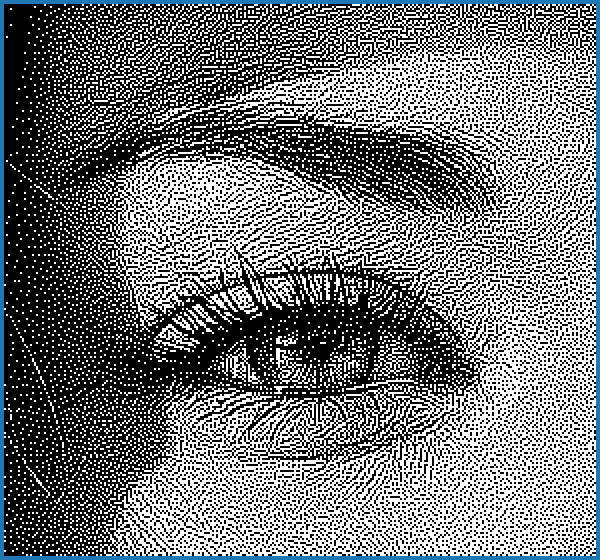}
      \vfill
      \vspace{0.5mm}
      \includegraphics[width=\linewidth]{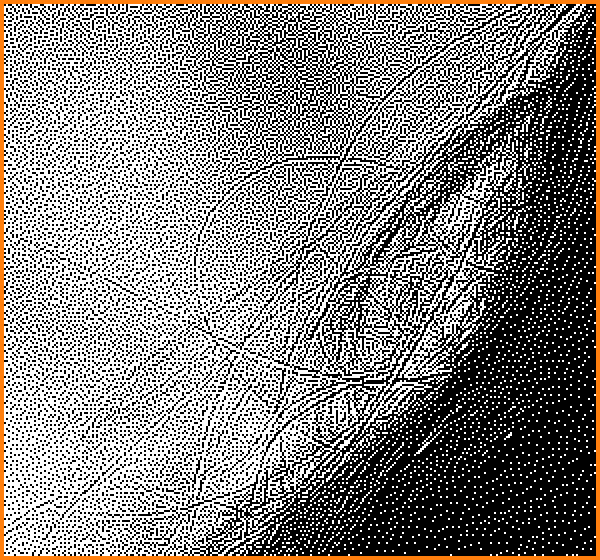}
      \vfill
      \small{SAED}
    \end{minipage}
    \hfill
    \begin{minipage}{0.24\linewidth}
      \centering
      \includegraphics[width=\linewidth]{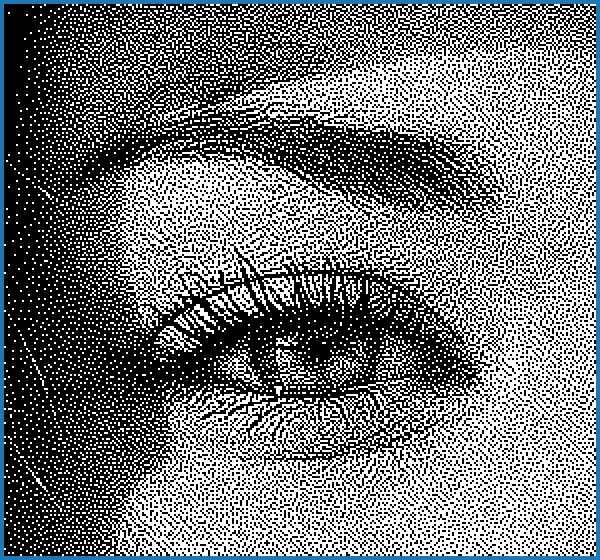}
      \vfill
      \vspace{0.5mm}
      \includegraphics[width=\linewidth]{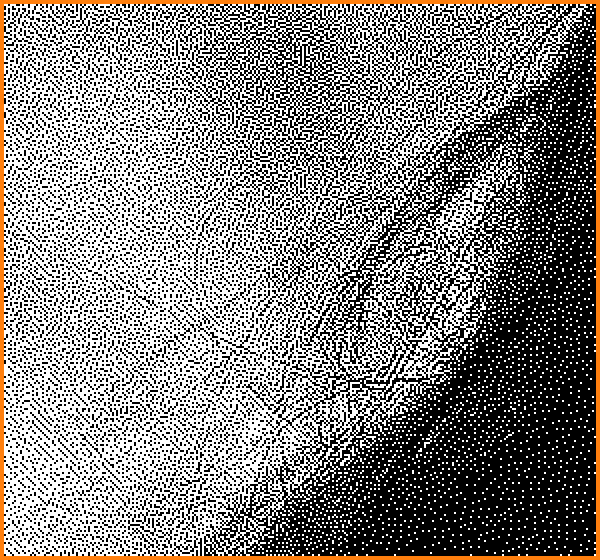}
      \vfill
      \small{cGAN}
    \end{minipage}
    \hfill
    \begin{minipage}{0.24\linewidth}
      \centering
      \includegraphics[width=\linewidth]{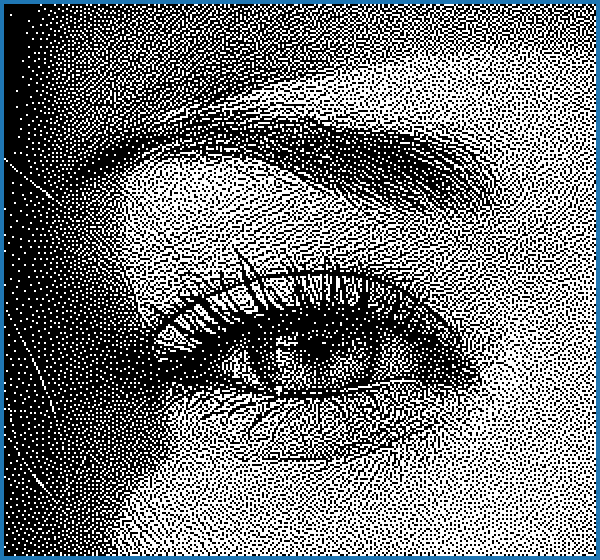}
      \vfill
      \vspace{0.5mm}
      \includegraphics[width=\linewidth]{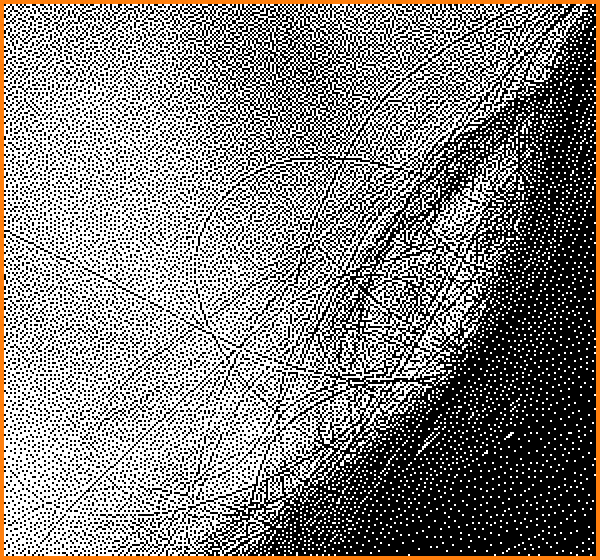}
      \vfill
      \small{Ours}
    \end{minipage}
  \end{minipage}
  \caption{Halftone samples of image ``Girl'' (from DIV2K dataset \cite{agustsson2017}). SAED and our method are capable of preserving fine structures such as the hairs in this example. The resolution of the full image is 2040x2040. SAED's runtime: 6,915 ms. cGAN's runtime: 2,703 ms. Our runtime: 483 ms.}
  \label{fig:div2k}
\end{figure*}

\textbf{Effect of $w_s$.}
The hyperparameter $w_s$ determines the contribution of structural similarity in the reward function.
To evaluate its effect, we plot the optimizing results of $\hat{g}_{\text{LE}}$ and $\hat{g}_{\text{COMA}}$ with different values (see Fig.~\ref{fig:w_s_sweep}).
In this work, we pick $w_s=0.06$ since the structure of its resulting halftone is clear enough without being overly sharp.
According to the quality preference, one can trade CSSIM for better PSNR by decreasing $w_s$, and vice versa.

\textbf{Effect of $\mathcal{L}_{AS}$ on Real Image.}
In the halftoning literature, the blue-noise property merits particular focus \cite{lau2008}.
For realistic images, we show two samples generated without $\mathcal{L}_{AS}$ (Fig.~\ref{fig:as_ablation}(a)) and with $\mathcal{L}_{AS}$ (Fig.~\ref{fig:as_ablation}(b)), as well as their Fourier amplitude spectra.
In Fig.~\ref{fig:as_ablation}(a), one can easily tell the regular distribution of dots with orientation preference, which brings peaks at certain phases on the spectrum.

\textbf{CNN Architecture.}
The proposed training framework can be applied to any CNN, so long as it keeps the image resolution.
We also train a UNet \cite{ronneberger2015} whose architecture strictly follows \cite{xia2021}'s practice.
The resultant PSNR (Näsänen/Gaussian), SSIM and CSSIM scores are 31.008/36.104/0.1580/0.9233, respectively.
First, we find that the UNet model is inferior to the ResNet in the halftoning task.
Second, the RL-based framework successfully activates the model to achieve higher metrics than RVH, while no additional halftone dataset or auxiliary NN is needed here.

\textbf{Comparison of Input Noise Type.}
In addition to the white Gaussian noise map $\mathcal{N}(0, 1)$ suggested by \cite{xia2021}, we have tested more common distributions including: uniform $\mathcal{U}(0, 1)$, gamma $\Gamma(1, 10)$ and Bernoulli $\mathrm{B}(0.5)$.
We also tested three special ``noise'' maps with spatial correlations: the position encoding proposed by \cite{kim2018}, the blue-noise dither array generated by \cite{ulichney1993}, and the DBS-screened halftones \cite{allebach1996} used in \cite{choi2022}.
Experiment results (see Fig.~\ref{fig:noise_type}) show that they achieve comparable quantitative quality.
For visual quality, the position encoding ``noise'' used in \cite{kim2018} can only generate periodic textures in the flat areas.
It would be an interesting topic to discuss how to render other halftone patterns \cite{frank2022} or other noise models \cite{fung2013}.

\textbf{Evaluation on DIV2K Dataset.}
In order to demonstrate the generalization ability of the proposed method, we also evaluated it on the DIV2K dataset \cite{agustsson2017}, which is split into 800 images for training and 100 images for testing.
The PSNR and CSSIM scores are 31.668 and 0.9334, respectively.
Fig.~\ref{fig:div2k} shows a sample from this dataset \cite{agustsson2017}.
We select SAED \cite{chang2009} and cGAN \cite{choi2022} (also retrained on DIV2K) for comparison here as the former shows ability to produce structural details, and the latter is currently the only published deep learning approach that focuses on generating aperiodic halftones.
One can see that our method can effectively capture details and demonstrate a significant increase in speed.
In addition, we tested the new model trained here on the VOC2012 \cite{voc2012} test set, and the PSNR/CSSIM scores are 31.535/0.9235, which is comparable to the original results (31.642/0.9236).

\subsection{An Example of Extending to Multitoning}

\begin{figure}[t]
  \centering
  \includegraphics[width=0.65\linewidth]{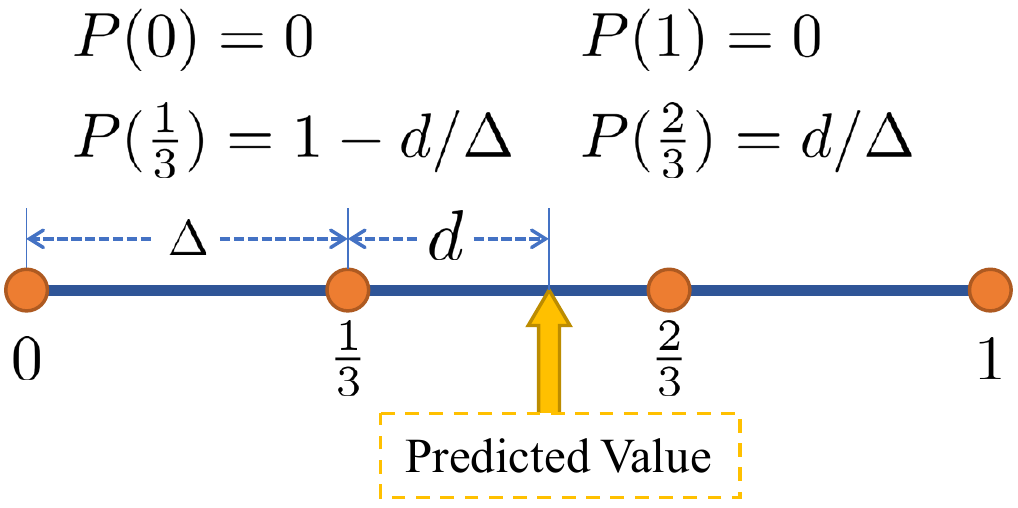}
  \caption{Illustration of casting CNN's outputs to the probabilities of multitone pixel's actions (levels=4).}
  \label{fig:multitone_cast}
\end{figure}

\begin{figure}[t]
  \centering
  \begin{minipage}{0.45\linewidth}
    \centering
    \includegraphics[width=\linewidth]{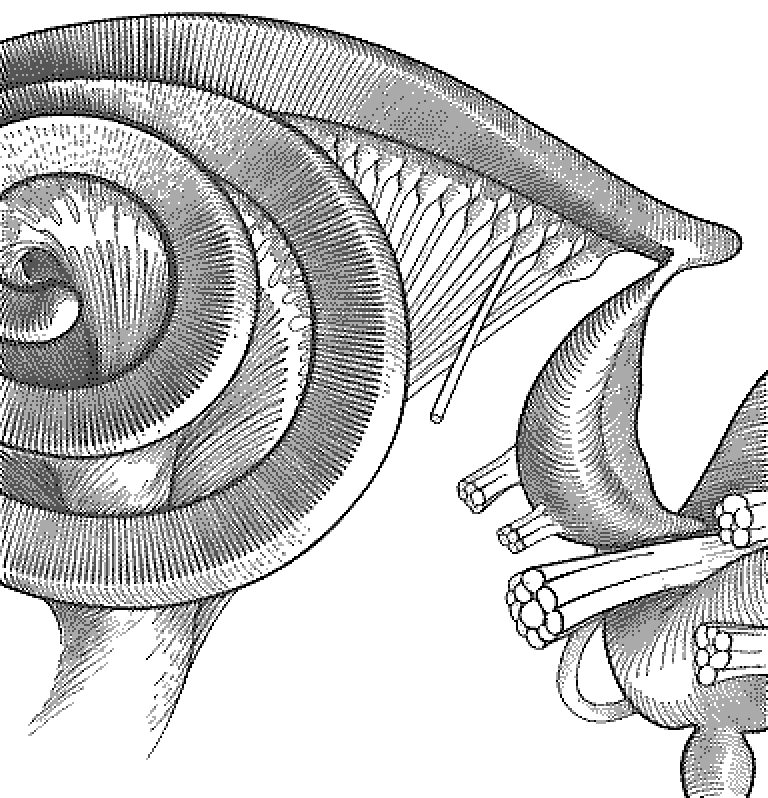}
  \end{minipage}
  \begin{minipage}{0.43\linewidth}
    \centering
    \includegraphics[width=\linewidth]{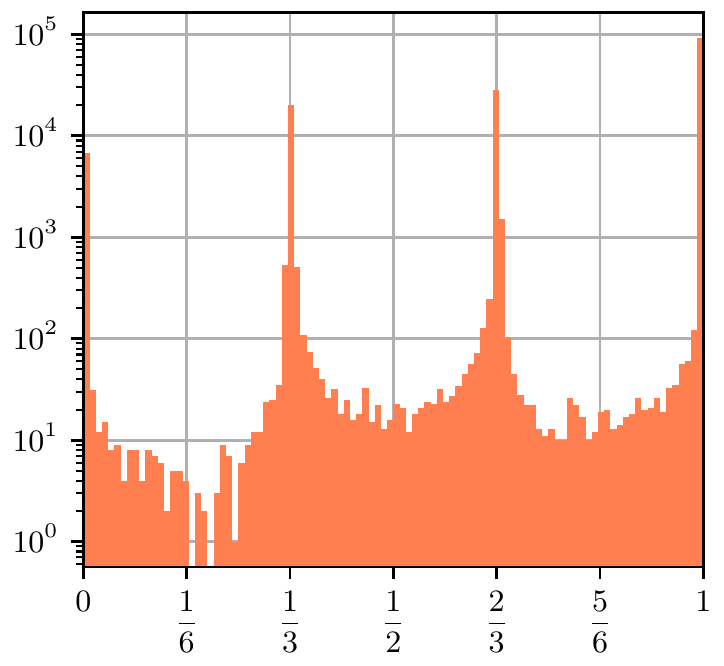}
  \end{minipage}
  \caption{Left: a multitone example generated by our deep multitoning model. Right: the histogram of the predicted values (note the logarithmic scale).}
  \label{fig:multitone}
\end{figure}

The framework presented above is exemplified by solving the basic binary halftoning problem, but it can be extended to more complex situations, such as color halftoning \cite{baqai2005} and multitoning \cite{rodriguez2008}.
To show the extensibility of our method, here we present a \emph{prototype} of deep multitoning, which reproduces a continuous-tone image with dots of more than two discrete ink intensities.
A straightforward solution is to let CNN generate multi-channel images by replacing the sigmoid layer with the softmax operation.
However, it not only brings a larger action space that we need to traverse (now the complexity is $O(N(L-1)+1)$, where $N$ is the pixel number and $L$ is the number of available ink levels), but also disables the directly applying of the proposed anisotropy estimation solution.
Motivated by some schemes from gradient quantization works \cite{zhou2016,alistarh2017}, we continue using the binary output model, but define the probabilities of actions as:
\begin{equation}
  P(h_a) =
  \begin{cases}
    (\text{ceil}(v_a) - v_a) / \Delta  & h_a=\text{floor}(v_a) \\
    (v_a - \text{floor}(v_a)) / \Delta & h_a=\text{ceil}(v_a)  \\
    0                                  & \text{otherwise},
  \end{cases}
\end{equation}
where $\text{ceil}(v_a)$ and $\text{floor}(v_a)$ denotes the nearest two discrete levels around $v_a$ predicted by the CNN, and $\Delta = 1/(L-1)$ is the distance between levels.
A 4-level case is illustrated in Fig.~\ref{fig:multitone_cast}.
The dimension of each agent's action space is still two, regardless of $L$, and we can directly extend the proposed algorithm to here with the same optimization target and training procedure.
A multitone sample and the predicted values' histogram (see Fig.~\ref{fig:multitone}) show that the majority of output values gather around the discrete levels.
As the number of available levels $L$ increases, finally the CNN learns an identical mapping.

Note that there are extra considerations in the multitoning problem, such as the different blue noise model \cite{rodriguez2008} and the banding artifact \cite{guo2015}.
While the extended solution presented here is computationally feasible, it can be further improved by taking these aspects into account.
It would not be trivial to conduct a systematical study of the multi-level situation.
Consequently, we leave them for future work.

\section{Conclusion}
\label{sec:conclu}
An efficient halftoning method with a data-driven methodology is proposed in this paper.
First, we propose to formulate halftoning as a reinforcement learning problem, in which an effective gradient estimator is tailored to train a light-weight CNN as the policy network.
Second, to achieve the blue-noise property, the anisotropy of constant grayscale images' halftone is suppressed by a new loss function in the training phase.
Finally, we suggest weighting the original SSIM metric by the contrast map of the input continuous-tone image to avoid the hole problem.
While existing halftoning methods use heuristics or conduct expensive search strategies, our trained model not only generates halftones with structural details but also stays efficient (15x faster than prior structure-aware method), as shown in the experiments.
Our framework can also be extended to address related problems such as multitoning.

We hope this work will motivate more colleagues to consider the potential benefits of deep learning for halftoning.
Future work could include:
\begin{itemize}
  \item A more domain-specific neural model to narrow the runtime gap between the present work and classic methods, such as error diffusion \cite{hu2016}.
  \item Extending to related problems, including, but not limited to, multitoning \cite{guo2015}, color halftoning \cite{agar2005}, video halftoning \cite{kao2022}, and other noise models \cite{fung2013,goyal2013}.
  \item Merging the deep halftoning model into real-time printer image processing pipelines.
\end{itemize}

\section*{Acknowledgments}
The authors would like to thank Aiguo~Yin from Pantum Electronics and Dr.~Menghan~Xia from Tencent AI Lab for their support in this work.

\bibliographystyle{IEEEtran}
\bibliography{refs}


\end{document}